\documentclass{article}

 \usepackage[final]{neurips_bdl2019}
 
 \usepackage[obeyFinal]{easy-todo}
     
\usepackage[utf8]{inputenc} % allow utf-8 input
\usepackage[T1]{fontenc}    % use 8-bit T1 fonts
\usepackage{hyperref}       % hyperlinks
\usepackage{url}            % simple URL typesetting
\usepackage{booktabs}       % professional-quality tables
\usepackage{amsfonts}       % blackboard math symbols
\usepackage{nicefrac}       % compact symbols for 1/2, etc.
\usepackage{microtype}      % microtypography

\usepackage{natbib}
\setcitestyle{square,numbers}

\usepackage{amsmath,graphicx}
\usepackage{amssymb}

\newcommand{\argmin}{\operatornamewithlimits{argmin}}

\def\qzx{q_{\boldsymbol \phi}({\bf z}|{\bf x})}
\def\qz{q_{\boldsymbol \phi}({\bf z})}
\def\pdx{p_{\mathcal D}({\bf x})}

\def\px_z{p({\bf x}|{\bf z})}
\def\pdc{p({\bf c})}
\def\pc_z{p({\bf c}|{\bf z})}
\def\pcz{p({\bf c},{\bf z})}
\def\pthetaz{p_{\boldsymbol \theta}({\bf z})}
\def\IXZ{I({\bf X}; {\bf Z})}
\def\IXZ_phi{I_{\boldsymbol \phi}({\bf X}; {\bf Z})}
\def\pzx{p({\bf z},{\bf x})}
\def\pz_x{p({\bf z}|{\bf x})}
\def\pthetax{p_{\boldsymbol \theta}({\bf x})}
\def\pthx_z{p_{\boldsymbol \theta}({\bf x}|{\bf z})}
\def\x{\bf x}
\def\c{\bf c}
\def\z{{\bf z}}

\def\X{\bf X}
\def\Z{\bf Z}
\def\C{\bf C}

\title{Information bottleneck through variational glasses}

\author{%
  Slava~Voloshynovskiy\thanks{Department of Computer Science,
University of Geneva, Carouge 1227, Switzerland}
  \\
  \texttt{svolos@unige.ch} \\
   \And
Mouad~Kondah$^*$
 \\
  \texttt{Mouad.Kondah@etu.unige.ch} \\
     \And
Shideh~Rezaeifar$^*$
 \\
  \texttt{Shideh.Rezaeifar@unige.ch} \\
       \And
Olga~Taran$^*$
 \\
  \texttt{Olga.Taran@unige.ch} \\
  \And
Taras~Holotyak$^*$
 \\
  \texttt{Taras.Holotyak@unige.ch} \\
       \And
Danilo~Jimenez~Rezende\thanks{DeepMind}
 \\
  \texttt{danilor@google.com} \\
}

\begin{document}

\maketitle

\vspace{-1mm}
\section{Abstract}

Information bottleneck (IB) principle \cite{tishby2015deep} has become an important element in  information-theoretic analysis of deep models. Many state-of-the-art generative models of both Variational Autoencoder (VAE) \cite{KingmaVAE,rezende2014stochastic} and Generative Adversarial Networks (GAN) \cite{goodfellow2014generative} families use various bounds on mutual information terms to introduce certain regularization constraints \cite{zhao2017infovae,  chen2016infogan, zhao2018information, alemi2018information, alemi2017fixing, alemi2016deep}. Accordingly, the main difference between these models consists in add regularization constraints and targeted objectives.

In this work, we will consider the IB framework for three classes of models that include supervised, unsupervised and adversarial generative models. We will apply a variational decomposition leading a common structure and allowing easily establish connections between these models and analyze underlying assumptions. 

Based on these results, we focus our analysis on unsupervised setup and reconsider the VAE family. In particular,  we present a new interpretation of  VAE family based on the IB framework using a direct decomposition of mutual information terms  and show some interesting  connections to existing methods such as VAE \cite{KingmaVAE,rezende2014stochastic}, $\beta-$VAE \cite{higgins2017beta}, AAE \cite{makhzani2015adversarial}, InfoVAE \cite{zhao2017infovae} and VAE/GAN \cite{larsen2015autoencoding}. Instead of adding regularization constraints to an evidence lower bound (ELBO) \cite{KingmaVAE,rezende2014stochastic}, which itself is a lower bound, we show that many known methods can be considered as a product of variational  decomposition of mutual information terms in the IB framework.
The proposed decomposition might also contribute to the interpretability of generative models of both VAE and GAN families and create a new insights to a generative compression \cite{agustsson2018generative, santurkar2018generative, tschannen2018deep, blau2019rethinking}. It can also be of interest for the analysis of novelty detection based on one-class classifiers \cite{pidhorskyi2018generative} with the IB based discriminators. 

{\bf Notations}: We will denote a joint generative distribution as $p_{\boldsymbol \theta}({\x},{\z}) =p_{\boldsymbol \theta}({\z}) p_{\boldsymbol \theta}({\x}|{\z})$, whereas marginal $p_{\boldsymbol \theta}({\z})$ is interpreted as a targeted distribution of latent space and marginal $p_{\boldsymbol \theta}({\x}) = \mathbb{E}_{p_{\boldsymbol \theta}({\z})}  \left[  p_{\boldsymbol \theta}({\x}|{\z})\right] = \int_{\z} p_{\boldsymbol \theta}({\x}|{\z})p_{\boldsymbol \theta}({\z}) \mathrm{d}{\z}$ as a generated data distribution with a generative model described by $p_{\boldsymbol \theta}({\x}|{\z})$. A joint  data distribution $q_{\boldsymbol \phi}({\x},{\z}) = \pdx q_{\boldsymbol \phi}({\z}|{\x})$, where $\pdx$ denotes an empirical data distribution and $q_{\boldsymbol \phi}({\z}|{\x})$ is an inference or encoding model and marginal $q_{\boldsymbol \phi}({\z})$ denotes a "true" or "aggregated" distribution of latent space data.

\section{Information bottleneck for different models}

In this section, we consider the IB framework and summarize some known results for supervised and unsupervised models. Having introduced a common base, we will also extend these results to generative adversarial models. Along this analysis, we will introduce several interesting bounds that will be used to develop a proposed bounded IB auto-encoding.

\subsection{Information bottleneck for supervised models}
\label{IB_supervised}

\begin{figure*}
    \centering
    \includegraphics[scale=0.75]{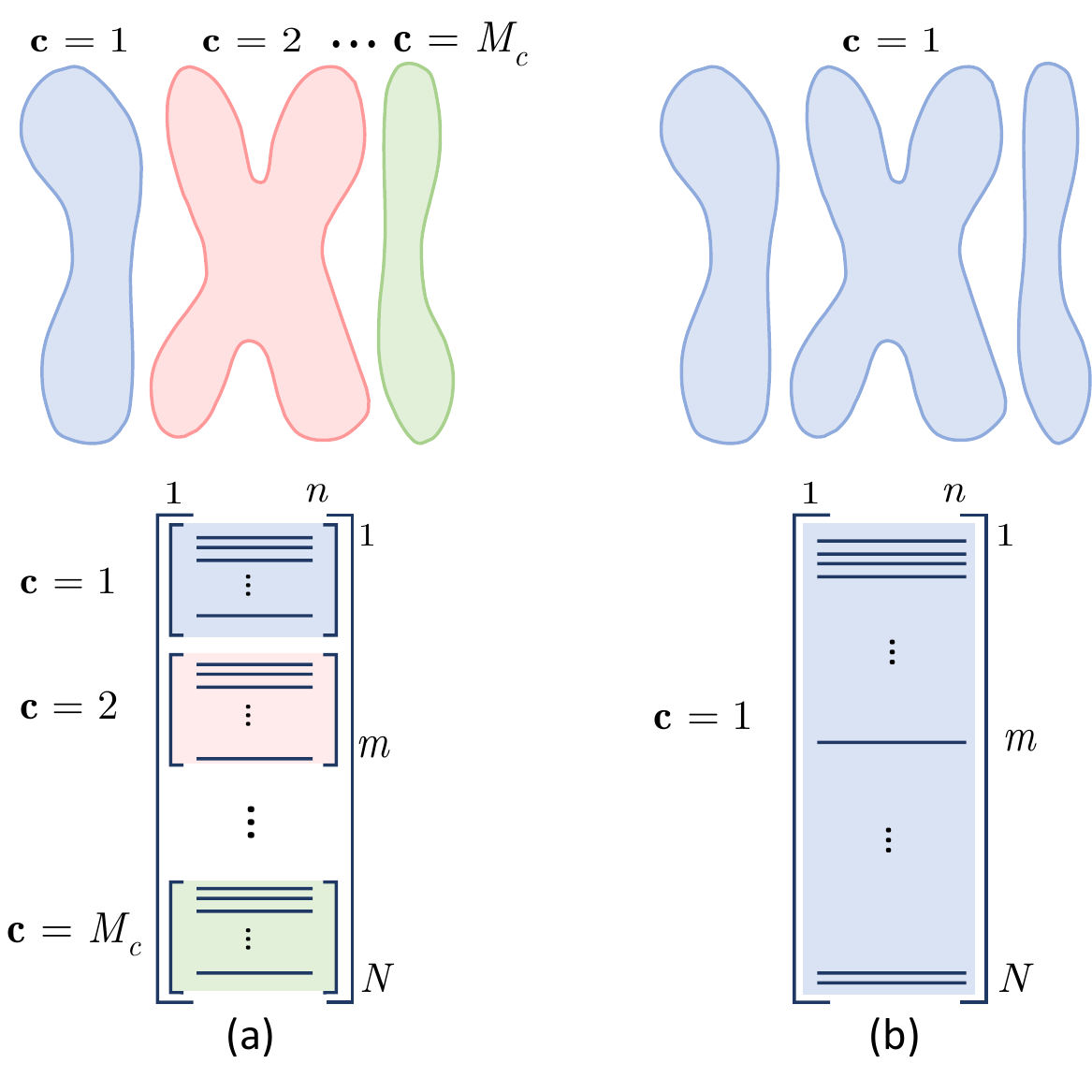}
    \caption{Labeling for supervised (a) and unsupervised (b) models. All data points
are shown as sequences of dimension $n$ at the bottom part of plot.}
    \label{Fig:labeling}
\end{figure*}

We consider a true joint distribution $p({\bf c},{\bf x})$ from which the training set $\{ {\bf x}_m, {\bf c}_m\}^{N}_{m=1}$ is sampled from, where each data sample is ${\bf x} \in \mathbb{R}^n$, $n$ denotes the dimensionality of data and $N$ stands for the number of training samples. We will use ${\bf c} \in \mathcal{M}$, with $ \mathcal{M} = \{1,\cdots,M_c\}$, to denote a class label. We use a vector notation for $\bf c$ to highlight that each label can be encoded according to some representation. The number of classes is denoted as $M_c$. The labeling of $N$ sequences into $M_c$ classes is shown in Figure \ref{Fig:labeling},a. It should be noted that many sequences might be assigned to the same class according to a set of chosen common features. We use different colors to reflect this labeling. At the same time, one can consider a "binning" organization principle shown in the bottom part of Figure \ref{Fig:labeling},a, where $N$ training sequences are allocated into $M_c$ bins representing $M_c$ classes.

\begin{figure*}
    \centering
    \includegraphics[scale=0.47]{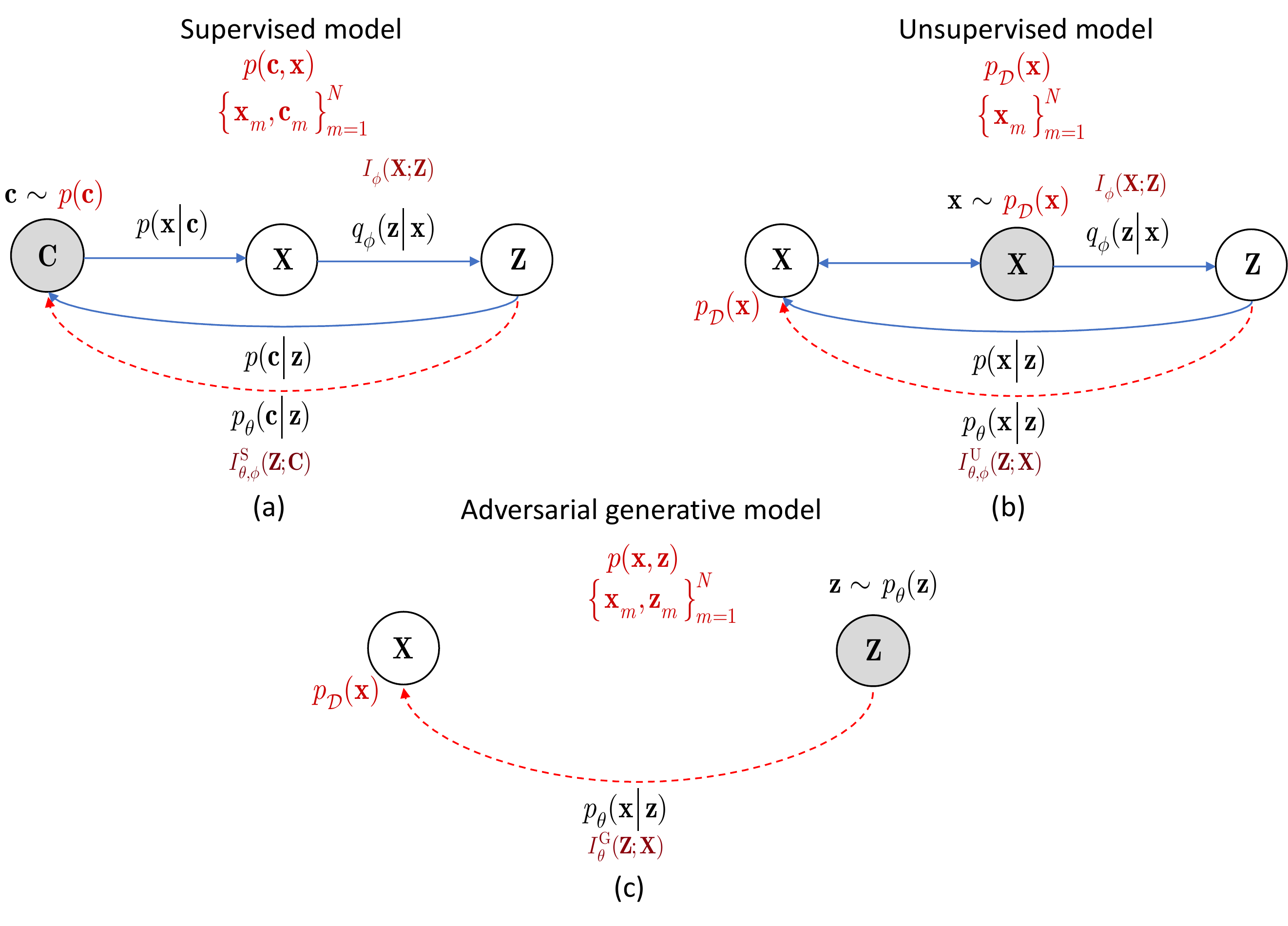}
    \caption{Information bottleneck models: (a) supervised, (b) unsupervised and (c) adversarial generative ones.}
    \label{Fig:models}
\end{figure*}

The supervised IB framework is considered based on Figure \ref{Fig:models},a. A sample $\bf x$ from a class $\bf c$ is generated by a mapping $p({\bf x}, {\bf c})=p({\bf c})p({\bf x}|{\bf c})$. The supervised IB can be formulated according to \cite{tishby2015deep} as:
\begin{equation}
\label{eq:IB_supervised}
\min_{\boldsymbol \phi: I({\bf Z}; {\bf C}) \geq I_c}  I_{\boldsymbol \phi}({\bf X};{\bf Z}).
\end{equation}
The supervised IB framework assumes an existence of a parametrized probabilistic mapping $q_{\boldsymbol \phi}(\mathbf{z} | \mathbf{x})$ with a controllable set of parameters $\boldsymbol \phi$, where $\z$ is considered to be a latent or bottleneck representation with dimensionality and statistical properties different of those of $\x$. It is assumed that three concerned vectors form a Markov chain ${\bf C} \rightarrow {\bf X} \rightarrow {\bf Z}$ and the objective is to find such a mapping $\boldsymbol \phi$, when $\z$ is a minimal sufficient statistic for task $\bf c$. The term $I_{\boldsymbol \phi}({\bf X};{\bf Z})$ denotes the mutual information between $\X$ and $\Z$ considering the above parametric mapping and $I({\bf Z}; {\bf C})$ corresponds to the mutual information between $\Z$ and $\C$.

The main idea behind the supervised IB (\ref{eq:IB_supervised}) consists in a search of parameters $\boldsymbol \phi$ that ensures the preservation of the information $I_c$ about the class $\bf c$ in the latent or bottleneck representation $\bf z$, while filtering out all irrelevant information from $\bf x$ that corresponds to the minimisation of $ I_{\boldsymbol \phi}({\bf X};{\bf Z})$ over $\boldsymbol \phi$. It should be pointed out that the minimization of mutual information can be obtained in different ways that include but are not limited to dimensionality reduction, compression that might include both clustering and quantization, additional of noise or sparsification of $\z$. All these techniques are well known and often used in practical deep net mappers implementing $q_{\boldsymbol \phi}(\mathbf{z} | \mathbf{x})$.

Tishby {\em et. al. } \cite{tishby2000information} also proposed the Langrangian of IB optimization (\ref{eq:IB_supervised}) defined as:
 \begin{equation} \label{eq:IB_supervised_Langangian_Tishby}
    \mathcal{L}^{\mathrm{S}}(\boldsymbol \phi) =  I_{\boldsymbol  \phi}({\bf  X} ; {\bf Z}) -\beta   I({\bf Z}; {\bf  C}),
  \end{equation}
where $\mathrm{S}$ stands for the supervised setup and $\beta$  is a regularization parameter corresponding to $I_c$ that leads to an optimization formulation:
 \begin{equation} \label{eq:IB_supervised_Langangian_Tishby_solution}
    {\boldsymbol {\hat \phi}} = \argmin_{\boldsymbol \phi}  
\mathcal{L}^{\mathrm{ S}}(\boldsymbol \phi).
  \end{equation}
  
In the following part, we will consider both terms of mutual information in (\ref{eq:IB_supervised_Langangian_Tishby}) and establish some useful bounds on them.

\subsubsection{Decomposition of the first term} 
 The first mutual information term $I_{\boldsymbol  \phi}({\bf  X} ; {\bf Z}) $ in (\ref{eq:IB_supervised_Langangian_Tishby}) is defined as:
 
 \begin{equation}
\label{MI_encoder_proof_definition}
\begin{aligned} 
 I_{\boldsymbol \phi}({\bf X} ;{\bf  Z}) & =  \mathbb{E}_{q_{\boldsymbol \phi}(\mathbf{z} , \mathbf{x}) } \left[  \log \frac{ q_{\boldsymbol \phi}(\mathbf{z} , \mathbf{x} )   }{ q_{\boldsymbol \phi}(\mathbf{z})p_{\mathcal D}(\mathbf{x})} \right]
  = \mathbb{E}_{q_{\boldsymbol \phi}(\mathbf{z} , \mathbf{x}) } \left[  \log \frac{ q_{\boldsymbol \phi}(\mathbf{z} | \mathbf{x} )  }{ q_{\boldsymbol \phi}(\mathbf{z})} \right]
 \\ & = H_{\boldsymbol \phi}({\bf Z}) - H_{\boldsymbol \phi}({\bf Z}|{\bf X}),
 \end{aligned} 
  \end{equation}
where $p_{\mathcal D}(\mathbf{x})$ denotes the data distribution and  $H_{\boldsymbol \phi}({\bf Z})  = - \mathbb{E}_{ q_{\boldsymbol \phi}(\mathbf{z}) } \left[\log  q_{\boldsymbol \phi}(\mathbf{z}) \right]$ denotes the entropy of distribution  $q_{\boldsymbol \phi}(\mathbf{z}) = \mathbb{E}_{p_{\mathcal D}(\mathbf{x}) } \left[ q_{\boldsymbol \phi}(\mathbf{z}|{\bf x}) \right]$ and $ H_{\boldsymbol \phi}({\bf Z}|{\bf X}) = - \mathbb{E}_{  q_{\boldsymbol \phi}(\mathbf{z},{\bf x}) } \left[ \log q_{\boldsymbol \phi}(\mathbf{z}|{\bf x}) \right]$ denotes the conditional entropy defined by $q_{\boldsymbol \phi}(\mathbf{z}|{\bf x})$. In (\ref{MI_encoder_proof_definition}), we used the decomposition of the joint distribution $q_{\boldsymbol \phi}(\mathbf{z},{\bf x}) = q_{\boldsymbol \phi}(\mathbf{z}|{\bf x}) p_{\mathcal D}(\mathbf{x})$. At the moment, we will not address technical details of computing $q_{\boldsymbol \phi}(\mathbf{z})$ and focus on them along the unsupervised setup analysis.
 
\subsubsection{Decomposition of the second term}  

The second mutual information term $I({\bf Z}; {\bf  C})$ in (\ref{eq:IB_supervised_Langangian_Tishby}) can be defined via $\pc_z$ as:
\begin{equation}
\label{MI_2nd_unsupervised_def1}
 I({\bf Z} ;{\bf  C})  = \mathbb{E}_{\pcz}  \left[  \log \frac{\pcz}{\pdc {p({\z})}}  \right] =\mathbb{E}_{\pcz}  \left[  \log \frac{\pc_z}{\pdc}  \right].
\end{equation}
We show in Appendix A, that this mutual information can be lower bounded by  $I({\bf Z};{\bf C}) \geq I^{\mathrm{S}}_{\boldsymbol \theta, \boldsymbol \phi}({\bf Z};{\bf C})$, where:
\begin{equation}
 \begin{aligned} 
\label{IB_S_lower_bound}
I^{\mathrm{S}}_{\boldsymbol \theta, \boldsymbol  \phi}({\bf Z};{\bf C}) & \triangleq  -  \mathbb{E}_{p({\bf c})}  \left[ \log p({\bf c}) \right] + \mathbb{E}_{p({\bf c},{\bf x} )}  \left[   \mathbb{E}_{q_{\boldsymbol \phi}({\bf z}|{\bf x})} \left[  \log p_{\boldsymbol \theta}(\mathbf{c}| \mathbf{z}) \right]\right]
 \\ & = H({\bf C}) - H_{\boldsymbol \theta, \boldsymbol \phi} ({\bf C}|{\bf Z}),
 \end{aligned} 
\end{equation}
with $H({\bf C})  = - \mathbb{E}_{p({\bf c})}  \left[ \log p({\bf c}) \right]$ and $H_{\boldsymbol \theta, \boldsymbol \phi} ({\bf C}|{\bf Z}) = - \mathbb{E}_{p({\bf c},{\bf x} )}  \left[   \mathbb{E}_{q_{\boldsymbol \phi}({\bf z}|{\bf x})} \left[  \log p_{\boldsymbol \theta}(\mathbf{c}| \mathbf{z}) \right]\right]$.

Therefore, the corresponding IB Lagrangian is redefined as:
 \begin{equation} \label{eq:IB_supervised_Langangian_bounded}
    \mathcal{L}^{\mathrm{S}}(\boldsymbol \phi, \boldsymbol \theta) =  I_{\boldsymbol  \phi}({\bf  X} ; {\bf Z}) -\beta   I^{\mathrm{S}}_{\boldsymbol \theta, \boldsymbol  \phi}({\bf Z};{\bf C}),
\end{equation}
that leads to the optimization problem:
 \begin{equation} 
 \label{eq:IB_supervised_optimization}
    ({\boldsymbol {\hat \theta},\boldsymbol {\hat \phi}}) = \argmin_{\boldsymbol {\theta}, \boldsymbol \phi}  
\mathcal{L}^{\mathrm{S}}(\boldsymbol \phi, \boldsymbol \theta).
  \end{equation}

{\bf Remark:} since $H({\bf C})$ in (\ref{IB_S_lower_bound}) is constant and does not depend on the parameters $\boldsymbol \theta, \boldsymbol \phi$, the supervised IB Lagrangian (\ref{eq:IB_supervised_Langangian_bounded}) can be rewritten in yet another commonly know form of supervised IB: 
 \begin{equation} 
 \label{eq:IB_supervised_Langangian_bounded_form2}
    \mathcal{L}^{\mathrm{S}}(\boldsymbol \phi, \boldsymbol \theta)  \propto  I_{\boldsymbol  \phi}({\bf  X} ; {\bf Z}) +\beta  H_{\boldsymbol \theta, \boldsymbol \phi} ({\bf C}|{\bf Z}).
\end{equation}
 In turns, it can be considered as finding a trade-off between the reduction of mutual information between $\X$ and $\Z$ according to the first term and the prediction accuracy of class $\c$ based on $\z$ according to the second term.

\subsection{Information bottleneck for unsupervised models}
\label{IB_unsupervised}
In the case of unsupervised setup, the data samples are not labelled by the classes $\bf c$. We will consider a true data distribution $p_{\mathcal D}({\bf x})$ from which the training set $\{ {\bf x}_m \}^{N}_{m=1}$ is sampled from. The data samples can be considered as belonging to a common class with the same label ${\bf c} = 1$ as shown in Figure \ref{Fig:labeling},b. Each sequence $\bf x$ is indexed by its proper index $m$. It means that the mapping between $m$ and $\bf x$ is unique $m \leftrightarrow {\bf x}$ in contrast to the supervised setup, where knowing $\bf c$ does not automatically imply that one knows a sample $\bf x$ but rather a set or bin to which it belongs to.

Alternatively, one can interpret the unsupervised setup as the supervised one with $M_c = N$ classes, i.e., when each class is represented by just one sequence as shown in Figure \ref{Fig:labeling}b. Therefore, by the direct analogy with the supervised setup, one can replace each class $\bf c$ by its proper representative sequence $\bf x$ as depicted in Figure \ref{Fig:models},b. Therefore, the generative process can be considered to start directly from $\bf x$ as shown by a gray circle. 

Thus, the unsupervised IB can be considered as a "compression" of $\bf x$ to $\bf z$ via the parametrized mapping $\qzx$ leading to a bottleneck representation $\bf z$ yet preserving a certain level of information $I_x$ in $\bf z$ about $\bf x$. Accordingly, the unsupervised IB problem can be formulated as: 
\begin{equation}
\label{eq:IB_unsupervised}
\min_{\boldsymbol \phi: I({\bf Z}; {\bf X}) \geq I_x}  I_{\boldsymbol \phi}({\bf X};{\bf Z}),
\end{equation}
and in the Lagrangian formulation as a minimization of:
 \begin{equation} \label{eq:IB_unsupervised_Langangian_Tishby}
    \mathcal{L}^{\mathrm{U}}(\boldsymbol \phi) =  I_{\boldsymbol  \phi}({\bf  X} ; {\bf Z}) -\beta   I({\bf Z}; {\bf  X}),
  \end{equation}
 where we use the same $\beta$ as for the supervised setup for the sake of simplicity and $\mathrm{U}$ denotes the unsupervised case.
 
 In the following sections, we will consider decompositions of both mutual information terms.
  
\subsubsection{Decomposition of the first term} 

The first term $\IXZ_phi$ in (\ref{eq:IB_unsupervised_Langangian_Tishby}) can be defined similarly to the supervised case (\ref{MI_encoder_proof_definition}) using entropies. The conditional entropy $H_{\boldsymbol \phi}({\bf Z}|{\bf X})$ is computable, since $\qzx$ is defined. However, the entropy $H_{\boldsymbol \phi}({\bf Z})  = - \mathbb{E}_{ q_{\boldsymbol \phi}(\mathbf{z}) } \left[\log  q_{\boldsymbol \phi}(\mathbf{z}) \right]$ requires computation of marginal distribution $q_{\boldsymbol \phi}(\mathbf{z}) = \mathbb{E}_{p_{\mathcal D}(\mathbf{x}) } \left[ q_{\boldsymbol \phi}(\mathbf{z}|{\bf x}) \right]$ that might be a computationally expensive task in practice. Therefore, we will proceed with a variational  approximation of $\qz$ by a distribution $\pthetaz$\footnote{Technically, the same factorization can be applied to the supervised counterpart (\ref{MI_encoder_proof_definition}). However, since in practice it is rarely of interest to generate labels $\bf c$ from $\z$, we only consider it in the scope of unsupervised generative and compression models.}:
\begin{equation}
\label{First_term_VA_unsupervised}
\begin{aligned} 
 I_{\boldsymbol \phi}({\bf X} ;{\bf  Z}) & =  \mathbb{E}_{q_{\boldsymbol \phi}(\mathbf{z} , \mathbf{x}) } \left[  \log \frac{ q_{\boldsymbol \phi}(\mathbf{x}, \mathbf{z} )}{ q_{\boldsymbol \phi}(\mathbf{z})p_{\mathcal{D}}(\mathbf{x})} \right] = 
   \mathbb{E}_{q_{\boldsymbol \phi}(\mathbf{z} , \mathbf{x}) } \left[  \log \frac{ q_{\boldsymbol \phi}(\mathbf{z} | \mathbf{x} )  }{ q_{\boldsymbol \phi}(\mathbf{z})} \frac{\pthetaz}{\pthetaz} \right]
 \\ & = \underbrace{\mathbb{E}_{p_{\mathcal D}(\mathbf{x})}\left[D_{\mathrm{KL}}\left(q_{\boldsymbol \phi}(\mathbf{z} | \mathbf{X=x}) \| p_{\boldsymbol \theta}(\mathbf{z} )\right) \right]}_\text{A}  -
  \underbrace{  D_{\mathrm{KL}}\left(q_{\boldsymbol \phi}(\mathbf{z} ) \| p_{\boldsymbol \theta}(\mathbf{z})\right)}_\text{B}, 
 \end{aligned} 
  \end{equation}
where the term (A) denotes the KL-divergence $\mathbb{E}_{p_{\mathcal D}(\mathbf{x})}\left[D_{\mathrm{KL}}\left(q_{\boldsymbol \phi}(\mathbf{z} | \mathbf{X=x}) \| p_{\boldsymbol \theta}(\mathbf{z} )\right) \right] = \mathbb{E}_{q_{\boldsymbol \phi}(\mathbf{z} , \mathbf{x}) } \left[  \log \frac{ q_{\boldsymbol \phi}(\mathbf{z} | \mathbf{x} )   }{ p_{\boldsymbol \theta}( \mathbf{z} ) }  \right]= \mathbb{E}_{p_{\mathcal D}(\mathbf{x})} \left[ \mathbb{E}_{q_{\boldsymbol \phi}(\mathbf{z} | \mathbf{x} )   }    \left[  \log \frac{ q_{\boldsymbol \phi}(\mathbf{z} | \mathbf{x} )   }{ p_{\boldsymbol \theta}( \mathbf{z} ) }  \right]  \right]$ and the term (B) denotes the KL-divergence $ D_{\mathrm{KL}}\left(q_{\boldsymbol \phi}(\mathbf{z} ) \| p_{\boldsymbol \theta}(\mathbf{z})\right) =   \mathbb{E}_{q_{\boldsymbol \phi}(\mathbf{z} , \mathbf{x}) } \left[  \log \frac{ q_{\boldsymbol \phi}(\mathbf{z}  )   }{ p_{\boldsymbol \theta}( \mathbf{z} ) }  \right]
 =   \mathbb{E}_{q_{\boldsymbol \phi}(\mathbf{z} ) } \left[  \log \frac{ q_{\boldsymbol \phi}(\mathbf{z}  )   }{ p_{\boldsymbol \theta}( \mathbf{z} ) }  \right]$.

\subsubsection{Decomposition of the second term} 
The second mutual information term $I({\bf Z}; {\bf  X})$ in (\ref{eq:IB_unsupervised_Langangian_Tishby}) is defined as:
\begin{equation}
\label{MI_2nd_unsupervised_def}
 I({\bf Z} ;{\bf  X})  =  \mathbb{E}_{\pzx}  \left[  \log \frac{\px_z}{\pdx}  \right].
\end{equation}

To find a variational approximation to the unknown $\px_z$, one can proceed in the same way as with the supervised model. However, one can also directly obtain a variational lower bound on $ I({\bf Z} ;{\bf  X})$ by assuming $\bf c \equiv \bf x$ in (\ref{IB_S_lower_bound}). This leads to  $I({\bf Z};{\bf X}) \geq I^{\mathrm{U}}_{\boldsymbol \theta, \phi}({\bf Z};{\bf X})$, where:
\begin{equation}
 \begin{aligned} 
\label{IB_U_lower_bound}
I^{\mathrm{U}}_{\boldsymbol \theta, \boldsymbol  \phi}({\bf Z};{\bf X}) & \triangleq 
  - \mathbb{E}_{\pdx} \left[ \log \pdx \right]
+ \mathbb{E}_{\pdx}  \left[   \mathbb{E}_{q_{\boldsymbol \phi}({\bf z}|{\bf x})} \left[  \log p_{\boldsymbol \theta}(\mathbf{x}| \mathbf{z}) \right]\right]
 \\ & = H_{\mathcal D}({\bf X}) - H_{\boldsymbol \theta, \boldsymbol \phi} ({\bf X}|{\bf Z}),
 \end{aligned} 
\end{equation}
with $H_{\mathcal D}({\bf X})  = - \mathbb{E}_{\pdx} \left[ \log \pdx \right]$ and $H_{\boldsymbol \theta, \boldsymbol \phi} ({\bf X}|{\bf Z}) = - \mathbb{E}_{\pdx}  \left[   \mathbb{E}_{q_{\boldsymbol \phi}({\bf z}|{\bf x})} \left[  \log p_{\boldsymbol \theta}(\mathbf{x}| \mathbf{z}) \right]\right]$.

Therefore, the corresponding IB Lagrangian is defined as:
 \begin{equation} \label{eq:IB_unsupervised_Langangian_bounded}
    \mathcal{L}^{\mathrm{U}}(\boldsymbol \phi, \boldsymbol \theta) =  I_{\boldsymbol  \phi}({\bf  X} ; {\bf Z}) -\beta   I^{\mathrm{U}}_{\boldsymbol \theta, \boldsymbol  \phi}({\bf Z};{\bf X}),
\end{equation}
thus leading to the minimization problem:
 \begin{equation} 
 \label{eq:IB_unsupervised_optimization}
    ({\boldsymbol {\hat \theta},\boldsymbol {\hat \phi}}) = \argmin_{\boldsymbol {\theta}, \boldsymbol \phi}  
\mathcal{L}^{\mathrm{U}}(\boldsymbol \phi, \boldsymbol \theta).
  \end{equation}

In should be pointed out that similarly to the supervised case (\ref{eq:IB_supervised_Langangian_bounded_form2}), the term $H_{\mathcal D}({\bf X})$ in (\ref{IB_U_lower_bound}) does not depend on the encoder and decoder parameters $\boldsymbol \phi, \boldsymbol \theta$ and can be skipped from the further consideration, if one is only concerned about the reconstruction task. 

Nevertheless, the same model can also be considered for a generative task, which will also be considered below, when a  trained encoder-decoder pair or just a sole decoder can be used for the generation of new samples from the latent space distribution. For these reasons, it is of interest to ensure that newly generated samples closely follow the statistics of original data. That is why one can also consider a decomposition of (\ref{IB_U_lower_bound}) as:
%\begin{equation}
% \begin{aligned} 
%\label{IB_U_lower_bound_KLD}
%I^{\mathrm{U}}_{\boldsymbol \theta, \boldsymbol  \phi}({\bf Z};{\bf X})   & =  \mathbb{E}_{\pdx} \left[ \mathbb{E}_{\pthetaz} \left[ \log \frac{\pthx_z}{\pdx} \right] \right]
% \\ & =  \mathbb{E}_{\pdx} \left[ \mathbb{E}_{\pthetaz} \left[ \log \frac{\pthx_z}{\pdx} \frac{\pthetax}{\pthetax} \right] \right]
% \\ & = - \mathbb{E}_{\pdx} \left[ \log \pthetax  \right] - \mathbb{E}_{\pdx} \left[ \log \frac{\pdx}{\pthetax} \right] + \mathbb{E}_{\pdx}  \left[   \mathbb{E}_{\pthetaz} \left[  \log p_{\boldsymbol \theta}(\mathbf{x}| \mathbf{z}) \right]\right]
%  \\ & = H(\pdx; \pthetax) - D_{\mathrm{KL}}\left(p_{\mathcal D}(\mathbf{x} ) \| p_{\boldsymbol \theta}(\mathbf{x})\right) + \mathbb{E}_{\pdx}  \left[   \mathbb{E}_{\pthetaz} \left[  \log p_{\boldsymbol \theta}(\mathbf{x}| \mathbf{z}) \right]\right].
% \end{aligned} 
%\end{equation} 
\begin{equation}
\hspace{-0.185cm}
 \begin{aligned} 
\label{IB_U_lower_bound_KLD}
I^{\mathrm{U}}_{\boldsymbol \theta, \boldsymbol  \phi}({\bf Z};{\bf X})   & =  \mathbb{E}_{\pdx} \left[ \mathbb{E}_{q_{\boldsymbol \phi}({\bf z}|{\bf x})} \left[ \log \frac{\pthx_z}{\pdx} \right] \right]
 \\ & =  \mathbb{E}_{\pdx} \left[ \mathbb{E}_{q_{\boldsymbol \phi}({\bf z}|{\bf x})} \left[ \log \frac{\pthx_z}{\pdx} \frac{\pthetax}{\pthetax} \right] \right]
 \\ & = - \mathbb{E}_{\pdx} \left[ \log \pthetax  \right] - \mathbb{E}_{\pdx} \left[ \log \frac{\pdx}{\pthetax} \right] + \mathbb{E}_{\pdx}  \left[   \mathbb{E}_{q_{\boldsymbol \phi}({\bf z}|{\bf x})} \left[  \log p_{\boldsymbol \theta}(\mathbf{x}| \mathbf{z}) \right]\right]
  \\ & = H(\pdx; \pthetax) - D_{\mathrm{KL}}\left(p_{\mathcal D}(\mathbf{x} ) \| p_{\boldsymbol \theta}(\mathbf{x})\right) + \mathbb{E}_{\pdx}  \left[   \mathbb{E}_{q_{\boldsymbol \phi}({\bf z}|{\bf x})} \left[  \log p_{\boldsymbol \theta}(\mathbf{x}| \mathbf{z}) \right]\right].
 \end{aligned} 
\end{equation} 
where $H(\pdx; \pthetax) = - \mathbb{E}_{\pdx} \left[ \log \pthetax  \right] $ denotes a cross-entropy.
Since $H(\pdx; \pthetax)  \geq 0$, one can lower bound (\ref{IB_U_lower_bound_KLD}) as $I^{\mathrm \bf {U}}_{\boldsymbol \theta, \boldsymbol  \phi}({\bf Z};{\bf X}) \geq  I^{\mathrm{U}_L}_{\boldsymbol \theta, \boldsymbol \phi}({\bf Z}; {\bf  X} )$, where\footnote{The cross-entropy computation requires knowledge of model $\pthetax$, whereas the KL-divergence is based on the ratio of two distributions and can be computed without an explicit knowledge of distributions but only from the training samples. For this reason, we proceed further with the KL-term.}:

\begin{equation}
\label{MI_decoder1}
\begin{aligned} 
 I^{\mathrm{U}_L}_{\boldsymbol \theta, \boldsymbol \phi}({\bf Z}; {\bf  X} )  & \triangleq   
 %+  \mathbb{E}_{p_{\mathcal D}({\bf x})}\left[D_{\mathrm{KL}}\left(q_{\boldsymbol \phi}(\mathbf{z} | \mathbf{x}) \| p_{\boldsymbol \theta}(\mathbf{z} | \mathbf{x})\right) \right] 
   \underbrace{  \mathbb{E}_{p_{\mathcal D}({\bf x})}\left[\mathbb{E}_{q_{\boldsymbol \phi}({\bf z} | {\bf x})}\left[\log p_{\boldsymbol \theta}({\bf x} | {\bf z})\right]\right]}_\text{C}
% - D_{\mathrm{KL}}\left(q_{\boldsymbol \phi}(\mathbf{z} ) \| p_{\boldsymbol \theta}(\mathbf{z})\right) 
-
 \underbrace{D_{\mathrm{KL}}\left(p_{\mathcal D}(\mathbf{x} ) \| p_{\boldsymbol \theta}(\mathbf{x})\right) }_\text{D}.
 %- 
  %\underbrace{\mathbb{E}_{ p_{\mathcal D}({\bf x})}\left[  \log p_{\mathcal D}(\mathbf{x}) \right] }_\text{E}.
 \end{aligned} 
\end{equation}

{\bf Remark:} The term (D) in (\ref{MI_decoder1}) can be implemented based on the density ratio estimation \cite{GoodfellowGAN} that will be addressed below. The term (C) can be defined explicitly using Gaussian or Laplacian priors. In the Laplacian case, one can define $p_{\boldsymbol \theta}(\mathbf{x}| \mathbf{z})  \propto \exp(-\lambda\| {\x} - g_{\boldsymbol \theta}({\z})\|_1)$ with a scale parameter $\lambda$, which leads to $\ell_1$-norm, and $g_{\boldsymbol \theta}({\z})$ denotes the decoder. It also corresponds to the model ${\x} = g_{\boldsymbol \theta}({\z}) + {\bf e}_x$, where ${\bf e}_x$ is a reconstruction error vector following the Laplacian pdf.
 Therefore, (\ref{MI_decoder1}) reduces to:
\begin{equation}
\label{MI_decoder11}
\begin{aligned} 
 I^{\mathrm{U}_L}_{\boldsymbol \theta, \boldsymbol \phi}({\bf Z}; {\bf  X} )  & = 
 %+  \mathbb{E}_{p_{\mathcal D}({\bf x})}\left[D_{\mathrm{KL}}\left(q_{\boldsymbol \phi}(\mathbf{z} | \mathbf{x}) \| p_{\boldsymbol \theta}(\mathbf{z} | \mathbf{x})\right) \right] 
   \underbrace{ - \lambda \mathbb{E}_{\pdx}  \left[   \mathbb{E}_{q_{\boldsymbol \phi}({\bf z} | {\bf x})} \left[ \| {\x} - g_{\boldsymbol \theta}({\z})\|_1) \right]\right]}_\text{C}
% - D_{\mathrm{KL}}\left(q_{\boldsymbol \phi}(\mathbf{z} ) \| p_{\boldsymbol \theta}(\mathbf{z})\right) 
-
 \underbrace{D_{\mathrm{KL}}\left(p_{\mathcal D}(\mathbf{x} ) \| p_{\boldsymbol \theta}(\mathbf{x})\right) }_\text{D}.
 %- 
  %\underbrace{\mathbb{E}_{ p_{\mathcal D}({\bf x})}\left[  \log p_{\mathcal D}(\mathbf{x}) \right] }_\text{E}.
 \end{aligned} 
\end{equation}

\subsubsection{Comparison of supervised and unsupervised IB}

Having considered the supervised and unsupervised IB formulations, it should be remarked several differences. 

The main origin of these differences is in the entropy of classes $H({\bf C})$ and entropy of data $H_{\mathcal D}({\X})$, i.e., $H_{\mathcal D}({\X}) \gg H({\bf C})$. The supervised IB describing the classification task only needs to ensure that the latent space data $\Z$, representing the sufficient statistics for $\C$, should preserve just $\log_2(M_c)$ bits to uniquely encode and recognize each class. In the unsupervised setup, the IB suggests to compress $\X$ to the encoding representation $\Z$ such that each sequence $\X$ is uniquely decodable or identifiable from $\Z$. It means that the entropy of latent space should correspond to the entropy of observation space, i.e., it should encode at least $\log_2 (N)$ bits to uniquely distinguish all $N$ sequences, unless some tolerance is allowed in terms of reconstruction error\footnote{The total number of samples in the training set is upper limited by $2^{nH(X)}$ under the i.i.d. assumption, whereas the training set is assumed to contain only $N$ sequences.}.

Naturally, this difference also leads to different encoding strategies. In the supervised setup, all common information within the same labeled class is "compressed" or disregarded and only the "differences" between the classes are encoded. With the increase of the number of classes, the differences might be minor that could be a potential source of vulnerability to adversarial attacks. An "informed" attacker knowing how these features are selected, that can be learned having an access to the same training data,  might change only several of them to achieve a flipping between the classes. In contrast, the entropy of latent data for the unsupervised setup should be considerably higher than those for the supervised setup. 

Finally, the nature of encoding is also different. In the unsupervised encoding, the classes are encoded to satisfy the reconstruction on average, i.e., the sequences close in the observation space might be close or even collude in the latent space, and the features of data contributing the most to the chosen metric of fidelity are preserved while less significant features are compressed or disregarded. As pointed above, all features that are irrelevant to a given classification task will be disregarded in the supervised setup. Using different re-labeling, new class-relevant features
will be extracted while class irrelevant information will be filtered out. In the unsupervised case, there is no labeling and the encoding solely depends on statistics of data.

\subsection{A link to generative adversarial models}
\label{IB_generative}

The generative adversarial models can be considered as in Figure \ref{Fig:models}c, i.e., the latent representation $\bf z$ of these models is not derived from the input of the network. Instead, it is assumed that the randomly assigned pairs $\{{\bf x}_m,{\bf z}_m\}_{m=1}^N$ are generated from $\pdx$ and $\pthetaz$.

Hence, the samples $\bf z$ are not produced by mapping $\pdx$ via $\qzx$ but directly from ${\z} \sim \pthetaz$ and thus the term $\IXZ_phi = 0$. Therefore, the unsupervised setup ({\ref{eq:IB_unsupervised_Langangian_bounded}}) reduces to the minimization of:
 \begin{equation} 
 \label{eq:IB_generative_Langangian_bounded}
   \hat{\boldsymbol \theta} = \min_{\boldsymbol \theta} \mathcal{L}^{\mathrm{G}}( \boldsymbol \theta),
\end{equation}
where $\mathcal{L}^{\mathrm{G}}( \boldsymbol \theta)= -\beta I^{\mathrm{G}}_{\boldsymbol \theta}({\bf Z};{\bf X})$ and:
\begin{equation}
 \begin{aligned} 
\label{IB_G_def}
I^{\mathrm{G}}_{\boldsymbol \theta}({\bf Z};{\bf X}) & \triangleq   \mathbb{E}_{\pdx} \left[ \mathbb{E}_{\pthetaz} \left[ \log \frac{\pthx_z}{\pdx} \right] \right],
 \end{aligned} 
\end{equation}
corresponds $ I^{\mathrm{U}}_{\boldsymbol \theta, \boldsymbol  \phi}({\bf Z};{\bf X})$ in ({\ref{eq:IB_unsupervised_Langangian_bounded}}) due to the fact that the sole link between $\Z$ and $\X$ is via $p_{\boldsymbol \theta}({\x}|{\z})$ and the latent vectors are generated from $\pthetaz$ and there is no dependence on $\boldsymbol \phi$.

Equivalently, the minimization problem (\ref{eq:IB_generative_Langangian_bounded}) can be reformulated as:
 \begin{equation} 
 \label{eq:IB_generative_Langangian_bounded_max}
    \hat{\boldsymbol \theta} = \max_{\boldsymbol \theta} I^{\mathrm{G}}_{\boldsymbol \theta}({\bf Z};{\bf X}). 
\end{equation}

Accordingly, using the factorization with respect to the marginal distribution of generated data $\pthetax$ similarly to the unsupervised case (\ref{IB_U_lower_bound_KLD}), one can define $I^{\mathrm{G}}_{\boldsymbol \theta}({\bf Z};{\bf X})$ as:

\begin{equation}
 \begin{aligned} 
\label{IB_G}
I^{\mathrm{G}}_{\boldsymbol \theta}({\bf Z};{\bf X}) & \triangleq   \mathbb{E}_{\pdx} \left[ \mathbb{E}_{\pthetaz} \left[ \log \frac{\pthx_z}{\pdx} \right] \right]
 \\ & =  \mathbb{E}_{\pdx} \left[ \mathbb{E}_{\pthetaz} \left[ \log \frac{\pthx_z}{\pdx} \frac{\pthetax}{\pthetax} \right] \right]
 \\ & = - \mathbb{E}_{\pdx} \left[ \log \pthetax  \right] - \mathbb{E}_{\pdx} \left[ \log \frac{\pdx}{\pthetax} \right] + \mathbb{E}_{\pdx}  \left[   \mathbb{E}_{\pthetaz} \left[  \log p_{\boldsymbol \theta}(\mathbf{x}| \mathbf{z}) \right]\right]
  \\ & = H(\pdx; \pthetax) - D_{\mathrm{KL}}\left(p_{\mathcal D}(\mathbf{x} ) \| p_{\boldsymbol \theta}(\mathbf{x})\right) + \mathbb{E}_{\pdx}  \left[   \mathbb{E}_{\pthetaz} \left[  \log p_{\boldsymbol \theta}(\mathbf{x}| \mathbf{z}) \right]\right].
 \end{aligned} 
\end{equation}

Since $H(\pdx; \pthetax)  \geq 0$, one can lower bound (\ref{IB_G}) as $I^{\mathrm{G}}_{\boldsymbol \theta}({\bf Z};{\bf X}) \geq I^{\mathrm{G}_L}_{\boldsymbol \theta}({\bf Z};{\bf X})$ where:
\begin{equation}
 \begin{aligned} 
\label{IB_G:lower_bound}
I^{\mathrm{G}_L}_{\boldsymbol \theta}({\bf Z};{\bf X}) \triangleq - D_{\mathrm{KL}}\left(p_{\mathcal D}(\mathbf{x} ) \| p_{\boldsymbol \theta}(\mathbf{x})\right) + \mathbb{E}_{\pdx}  \left[   \mathbb{E}_{\pthetaz} \left[  \log p_{\boldsymbol \theta}(\mathbf{x}| \mathbf{z}) \right]\right].
 \end{aligned} 
\end{equation}

Similarly to (\ref{MI_decoder11}), one can further develop (\ref{IB_G:lower_bound}) using $ p_{\boldsymbol \theta}(\mathbf{x}| \mathbf{z})  \propto \exp(-\lambda\| {\x} - g_{\boldsymbol \theta}({\z})\|_1)$ with a scale parameter $\lambda$ that results in:

%{\bf Remark:} the lower bound $I^{\mathrm{G}_L}_{\boldsymbol \theta}({\bf Z};{\bf X})$ (\ref{IB_G:lower_bound}) contains two terms. The first term can be implemented based on the density ratio estimation. The send term can be defined explicitly using Gaussian or Laplacian priors.
%
%$ p_{\boldsymbol \theta}(\mathbf{x}| \mathbf{z})  \propto \exp(-\lambda\| {\x} - g_{\boldsymbol \theta}({\z})\|_1)$ with a scale parameter $\lambda$, which leads to $\ell_1$-norm. 
%

\begin{equation}
 \begin{aligned} 
\label{IB_G:lower_bound_final}
   \hat{\boldsymbol \theta}  = \max_{\boldsymbol \theta} I^{\mathrm{G}_L}_{\boldsymbol \theta}({\bf Z};{\bf X}) =   \min_{\boldsymbol \theta}  D_{\mathrm{KL}}\left(p_{\mathcal D}(\mathbf{x} ) \| p_{\boldsymbol \theta}(\mathbf{x})\right) + \lambda \mathbb{E}_{\pdx}  \left[   \mathbb{E}_{\pthetaz} \left[ \| {\x} - g_{\boldsymbol \theta}({\z})\|_1) \right]\right].
 \end{aligned} 
\end{equation}

{\bf Remark :} Vanilla GANs use only an approximation to the first term for the generator optimization. However, GANs might face a mode collapse and the likelihood term can at least theoretically regularize it.

\section{Bounded information bottleneck AE formulation}
\vspace{-1mm}
Having considered the unsupervised and adversarial generative models, we can proceed with the formulation of a new auto-encoding framework. More particularly, we will use the results (\ref{First_term_VA_unsupervised}) and (\ref{MI_decoder1}) to propose a new type of unsupervised auto-encoder that combines the elements of VAE and GAN families and is built on the IB principle. We will refer to this auto-encoder as a {\em bounded information bottleneck AE} (BIB-AE) and link it to the VAE family of auto-encoders, generative compression and one-class classification. It should also be pointed out that the BIB-AE framework is rather considered as a conceptual generalization then as practical implementation. However, we will comment how to implement the BIB-AE components in practice using known techniques of KL-divergence approximation.

The BIB-AE Lagrangian is based on (\ref{eq:IB_unsupervised_Langangian_bounded}) and is defined as:
\begin{equation}
\label{BIBN_unsupervised}
\mathcal{L}_{\mathrm{BIB-AE}}(\boldsymbol \theta, \boldsymbol \phi) =  I_{\boldsymbol  \phi}({\bf  X} ; {\bf Z}) -\beta    I^{\mathrm{U}_L}_{\boldsymbol \theta, \boldsymbol \phi}({\bf Z}; {\bf  X} ),
 \end{equation}
where $I_{\boldsymbol  \phi}({\bf  X} ; {\bf Z})$ and $I^{\mathrm{U}_L}_{\boldsymbol \theta, \boldsymbol \phi}({\bf Z}; {\bf  X} )$ correspond to (\ref{First_term_VA_unsupervised}) and (\ref{MI_decoder1}) that we summarize below for the convenience of analysis: 
\begin{equation}
\label{MI_encoder_sum}
\begin{aligned} 
 I_{\boldsymbol \phi}({\bf X} ;{\bf  Z}) & =  \underbrace{\mathbb{E}_{p_{\mathcal D}(\mathbf{x})}\left[D_{\mathrm{KL}}\left(q_{\boldsymbol \phi}(\mathbf{z} | \mathbf{X=x}) \| p_{\boldsymbol \theta}(\mathbf{z} )\right) \right]}_\text{A}  -
  \underbrace{  D_{\mathrm{KL}}\left(q_{\boldsymbol \phi}(\mathbf{z} ) \| p_{\boldsymbol \theta}(\mathbf{z})\right)}_\text{B}, 
  \end{aligned} 
\end{equation}
\begin{equation}
\label{MI_decoder_sum}
\begin{aligned} 
I^{\mathrm{U}_L}_{\boldsymbol \theta, \boldsymbol \phi}({\bf Z}; {\bf  X} ) & =   
 %+  \mathbb{E}_{p_{\mathcal D}({\bf x})}\left[D_{\mathrm{KL}}\left(q_{\boldsymbol \phi}(\mathbf{z} | \mathbf{x}) \| p_{\boldsymbol \theta}(\mathbf{z} | \mathbf{x})\right) \right] 
   \underbrace{  \mathbb{E}_{p_{\mathcal D}({\bf x})}\left[\mathbb{E}_{q_{\boldsymbol \phi}({\bf z} | {\bf x})}\left[\log p_{\boldsymbol \theta}({\bf x} | {\bf z})\right]\right]}_\text{C}
% - D_{\mathrm{KL}}\left(q_{\boldsymbol \phi}(\mathbf{z} ) \| p_{\boldsymbol \theta}(\mathbf{z})\right) 
-
 \underbrace{D_{\mathrm{KL}}\left(p_{\mathcal D}(\mathbf{x} ) \| p_{\boldsymbol \theta}(\mathbf{x})\right) }_\text{D}.
 %- 
  %\underbrace{\mathbb{E}_{ p_{\mathcal D}({\bf x})}\left[  \log p_{\mathcal D}(\mathbf{x}) \right] }_\text{E}.
 \end{aligned} 
\end{equation}

The BIB-AE parameters are found according to the following minimization problem:
 \begin{equation} 
 \label{eq:BIB_optimization}
    ({\boldsymbol {\hat \theta},\boldsymbol {\hat \phi}}) = \argmin_{\boldsymbol {\theta}, \boldsymbol \phi}  
\mathcal{L}_{\mathrm{BIB-AE}}(\boldsymbol \theta, \boldsymbol \phi).
  \end{equation}

\begin{figure*}
    \centering
    \includegraphics[scale=0.47]{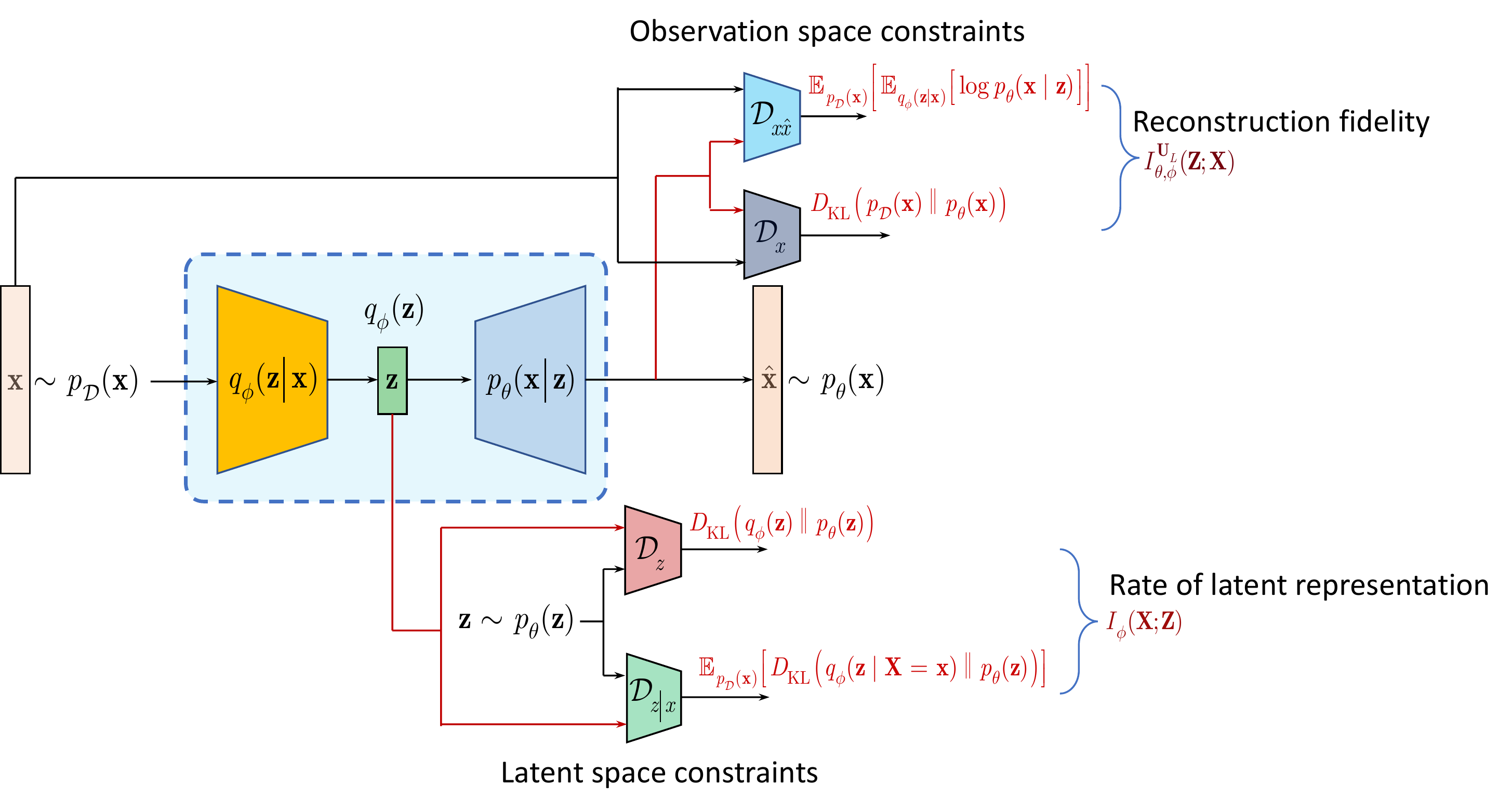}
    \caption{Generalized diagram of BIB-AE.}
    \label{Fig:set_up}
\end{figure*}

The diagram explaining the BIB-AE setup is shown in Figure \ref{Fig:set_up}. The reconstruction fidelity is ensured jointly by the terms (C) and (D), while the minimization of mutual information between $\X$ and $\Z$ is guided by the targeted distribution of the latent space $p_{\boldsymbol \theta}({\bf z})$ according to the terms (A) and (B). The "stochasticity" of the encoder will determine to which extend the mappings of data points from the observation space will "overlap" in the latent space yet satisfying the correspondence between the marginal posterior and the prior.    

%The minimization of $ I_{\boldsymbol \phi}({\bf X} ;{\bf  Z}) $ is seen as finding such an encoder mapping $q_{\boldsymbol \phi}(\mathbf{z} | \mathbf{x})$ that would minimize the difference between the terms (A) and (B) that defines the correspondence between conditional $q_{\boldsymbol \phi}(\mathbf{z} | \mathbf{x})$ and marginal $q_{\boldsymbol \phi}(\mathbf{z})$ distributions with respect the targeted $\pthetaz$. 

%not a sole objective to achieve a reduction of mutual information but also to ensure that both the conditional distribution $q_{\boldsymbol \phi}({\bf z}|{\bf x})$ and marginal distribution $q_{\boldsymbol \phi}({\bf z})$ match  the targeted distribution $p_{\boldsymbol \theta}({\bf z})$. 

\begin{figure*}
    \centering
    \includegraphics[scale=0.47]{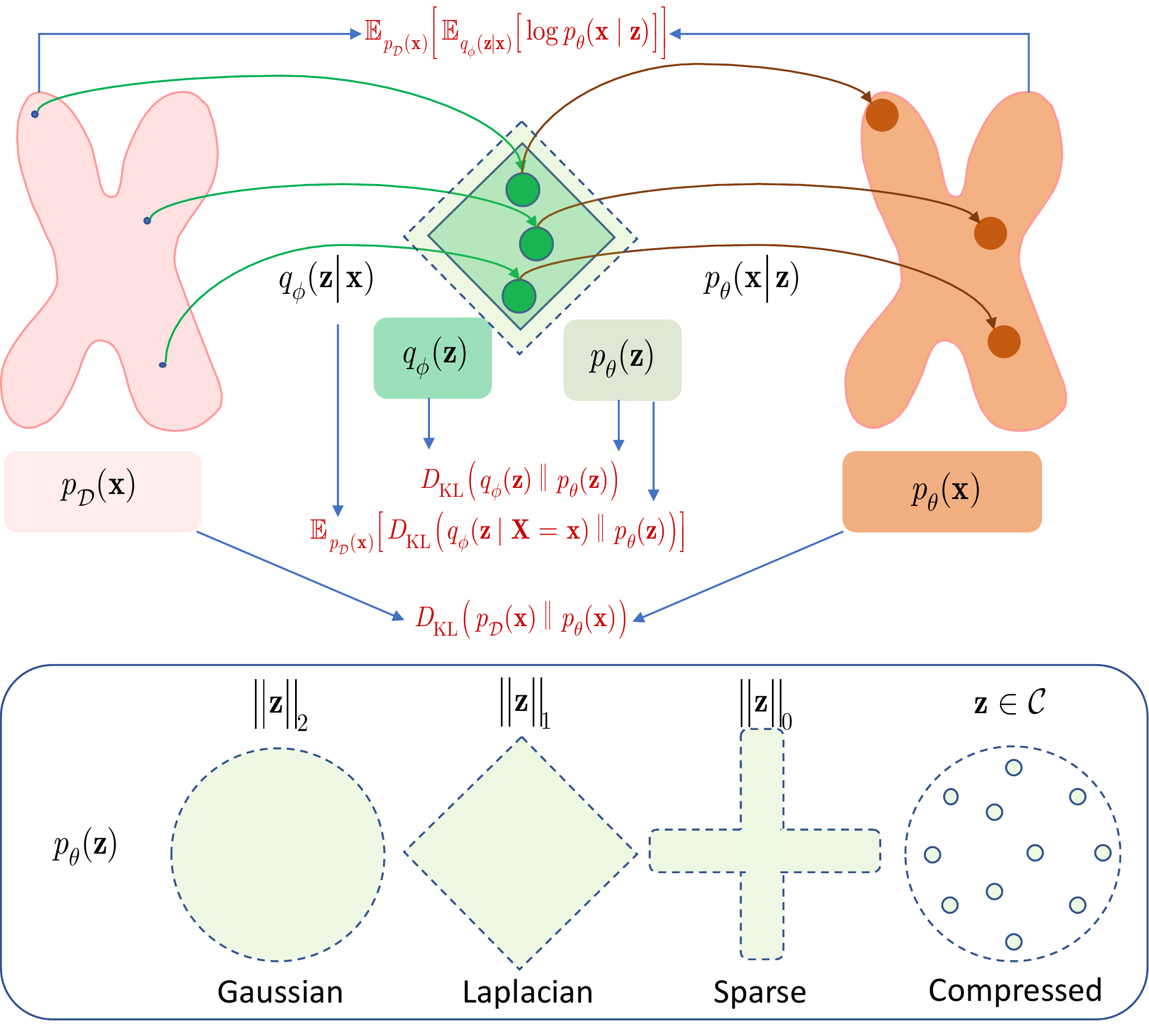}
    \caption{BIB-AE as a stochastic mapping. Possible targeted priors of latent space are shown in the bottom of figure.}
    \label{Fig:3}
\end{figure*}

More particularly, as shown in Figure \ref{Fig:3}, the data distribution $\pdx$ is mapped to the latent space marginal  distribution $\qz$ via the stochastic mapping $\qzx$. According to the variational approach, the targeted distribution of latent space is $\pthetaz$ and the encoder tries to optimize the parameters of encoder $\boldsymbol \phi$ according to (\ref{eq:BIB_optimization}) to meet both the constraints on the latent space and the reconstruction fidelity by satisfying the targeted $\beta    I^{\mathrm{U}_L}_{\boldsymbol \theta, \boldsymbol \phi}({\bf Z}; {\bf  X} ) $. One can imagine several forms of stochastic encoding: (i) ${\z} = f_{\boldsymbol \phi}({\x}) + {\boldsymbol \epsilon}$, where ${\boldsymbol \epsilon}$ follows the distribution defying the properties  of conditional distribution $\qzx$, (ii) ${\z} = f_{\boldsymbol \phi}({\x} + {\boldsymbol \epsilon})$ or (iii) ${\z} = f_{\boldsymbol \phi}( [{\x}, {\boldsymbol \epsilon}])$.
However, in practice depending on a chosen way of computing KL-divergence, one might be interested in a tractable density. In this case, the encoding of the first type is used as for example in the VAE family. Disregarding a particular form of randomness injecting mechanism, the green circles in the latent space of Figure \ref{Fig:3}  denote the resulting stochastic mappings of each point from the observable space. 

%%%%%{\bf TO consider several cases: the exact value of MI is needed or not, if we try to estimate it directly; and instead of computing the KL divergence one can try to estimate it via various tricks: see above.}

%ensure that $\qz$ matches $\pthetaz$ yet meeting the conflicting constraint that simultaneously  

%\todo{To describe Fig 5}

  \begin{figure*}
    \centering
    \includegraphics[scale=0.45]{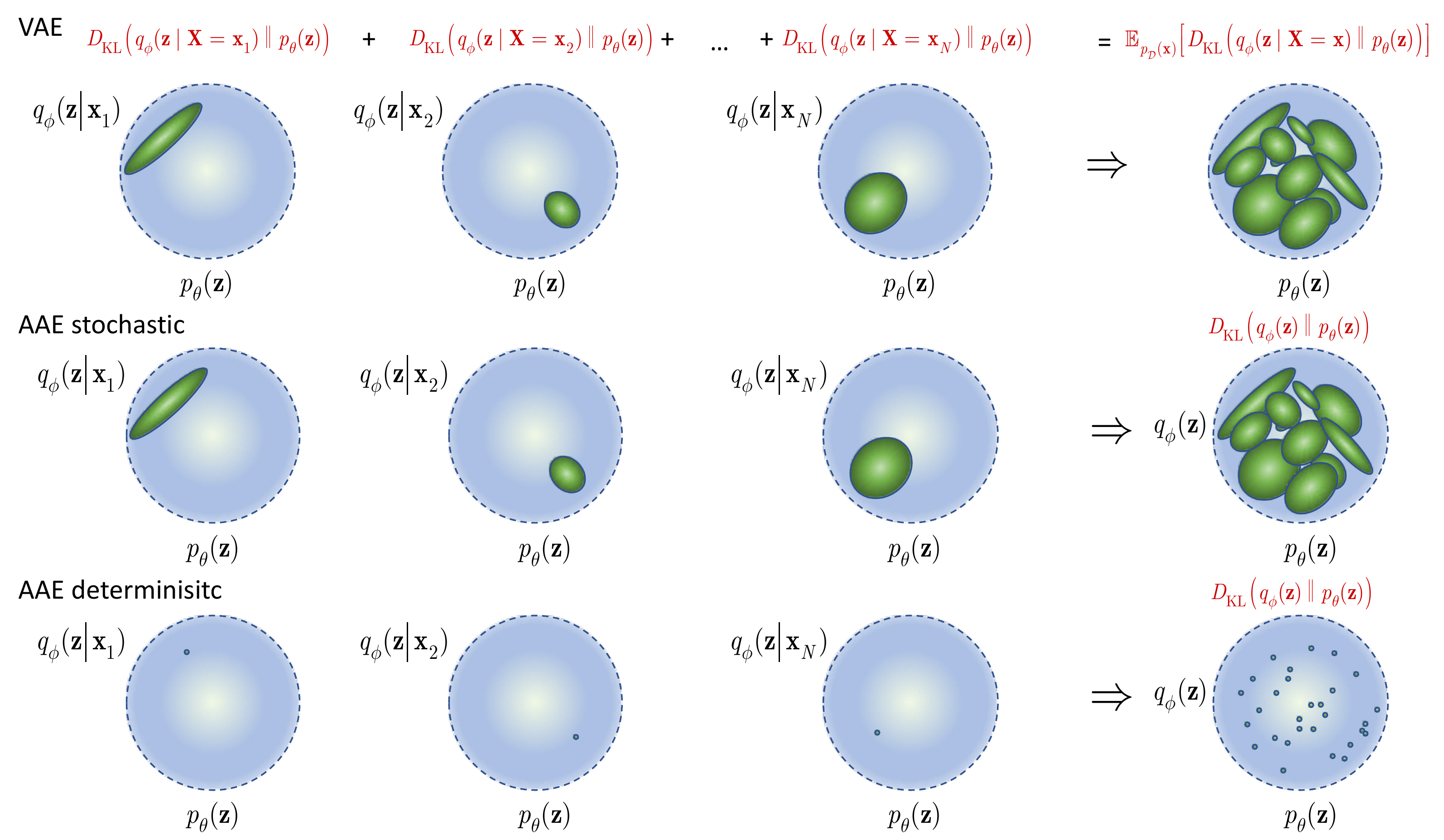}
    \caption{Schematic visualization of latent space for VAE and AAE.}
    \label{Fig:4}
\end{figure*}

%The training of the system is performed by an alternating procedure when\footnote{The encoder and decoder can be pre-trained in the normal AE regime, i.e., by the minimization of the term (C) only. }: (i) first the encoder and the observation space discriminator $D_x$, defined by the term (D), are trained, (ii) then the encoder and the latent space "discriminators" $D_z$, defined by the term (B), and $D_{z|x}$, defined by the term (A), are trained  and finally (iii) the encoder and decoder are re-trained to minimize the term (C).

%{\bf Remark 1}:

%note on the architecture: it combines different schemes considered above: (i) at the first stage one can either train the classical AE to gent an estimate of encoder and decoder or alternatively train only the decoder in the described GAN mode using MSE and Dx; (ii) then use any guess or pre-trained encoder to satisfy the constraints on the latent space and then (iii) optimize both encoder and decoder for this latent space representation.

%\todo{1. To add images of all AE and compressors as in slides. 2. To explain that VAE pushes condt pdf to be close to targeted marginal pdf, i.e., it pushes it to be zero mean. So all samples are mapped almost to the same region.}

\vspace{-1mm}
\section{Connections to the prior art AEs}
\vspace{-1mm}
\subsection{Generative models of VAE family}
\vspace{-1.2mm}
{\bf VAE \cite{KingmaVAE,rezende2014stochastic}} Lagrangian is defined as:
\begin{equation}
\label{VAE}
\mathcal{L}_{\mathrm{VAE}}(\boldsymbol \theta, \boldsymbol \phi) = \mathbb{E}_{p_{\mathcal D}({\bf x})}\left[  D_{\mathrm{KL}}(q_{\boldsymbol \phi}(\mathbf{z}|{\bf X= x} ) \| p_{\boldsymbol \theta}(\mathbf{z}))\right] - \mathbb{E}_{p_{\mathcal D}({\bf x})}\left[\mathbb{E}_{q_{\boldsymbol \phi}({\bf z} | {\bf x})}\left[\log p_{\boldsymbol \theta}({\bf x} | {\bf z})\right]\right],
 \end{equation}
and contains only 2 terms (A) and (C) in (\ref{BIBN_unsupervised}) with $ \beta =1$. It can be shown that the VAE is based on an upper bound on $ I_{\boldsymbol \phi}({\bf X} ;{\bf  Z}) \leq  I^{U}_{\boldsymbol \phi}({\bf X} ;{\bf  Z}) =  \mathbb{E}_{p_{\mathcal D}(\mathbf{x})}\left[D_{\mathrm{KL}}\left(q_{\boldsymbol \phi}(\mathbf{z} | \mathbf{X=x}) \| p_{\boldsymbol \theta}(\mathbf{z} )\right) \right]$, since $ 
  D_{\mathrm{KL}}\left(q_{\boldsymbol \phi}(\mathbf{z} ) \| p_{\boldsymbol \theta}(\mathbf{z})\right) \geq 0$.  Similarly, since $D_{\mathrm{KL}}\left(p_{\mathcal D}(\mathbf{x} ) \| p_{\boldsymbol \theta}(\mathbf{x})\right) \geq 0$, and denoting $I^{\text{VAE}} _{\boldsymbol \theta, \boldsymbol \phi}({\bf Z}; {\bf  X} ) = \mathbb{E}_{p_{\mathcal D}({\bf x})}\left[\mathbb{E}_{q_{\boldsymbol \phi}({\bf z} | {\bf x})}\left[\log p_{\boldsymbol \theta}({\bf x} | {\bf z})\right]\right]$, one obtains   
   $ I^{\text{VAE}} _{\boldsymbol \theta, \boldsymbol \phi}({\bf Z}; {\bf  X} )  \geq  I^{\mathrm{U}_L}_{\boldsymbol \theta, \boldsymbol \phi}({\bf Z}; {\bf  X} )$.
 
 The VAE encoder can be considered as a stochastic mapping with a particular form of parametrization \cite{KingmaVAE} ${\z} = {\boldsymbol \mu}({\bf x}) + {\boldsymbol \sigma}({\bf x})\odot{\boldsymbol \epsilon}$, where ${\boldsymbol \mu}({\bf x})$  and ${\boldsymbol \sigma}({\bf x})$ are outputs of the network $f_{\boldsymbol \phi} ({\bf x})$ and ${\boldsymbol \epsilon}$ is assumed to be a zero mean unit variance vector, i.e.,  ${\boldsymbol \epsilon} \sim \mathcal{N}({\bf 0},{\bf I})$, and $\odot$ denotes element wise product. As a result, the conditional distribution of $\Z$ given an input variable $\X$ follows a Gaussian distribution $q_{\boldsymbol \phi} ({\z}|{\x}) = \mathcal{N}({\boldsymbol \mu}({\bf x}), \text{diag}({\boldsymbol \sigma}({\bf x})))$. The VAE also assumes a prior on the latent space to be $p_{\boldsymbol \theta}({\z}) = \mathcal{N}({\bf 0},{\bf I})$. Under these conditions the KL-term (A) can be computed analytically. 
 
 %At the same time, one can choose different priors on the latent space distribution as shown in Figure \ref{Fig:3}.

It should be pointed out that the VAE encoder maps a point from the observation space into a probabilistic output of Gaussian cloud with mean ${\boldsymbol \mu}({\bf x})$ and "ellipsoid" orientation determined by the diagonal covariance matrix $\text{diag}({\boldsymbol \sigma}({\bf x}))$. This is schematically shown in a form of green ellipsoids for different samples ${\x}_i$, $i=1,\cdots, N$, in the latent space according to Figure \ref{Fig:4}.  Moreover, since the targeted marginal prior is  $p_{\boldsymbol \theta}({\z}) = \mathcal{N}({\bf 0},{\bf I})$, and the KL term for all mappings of ${\x}$'s via $q_{\boldsymbol \phi}(\mathbf{z} | \mathbf{x})$ should match this $p_{\boldsymbol \theta}({\z})$ in $\mathbb{E}_{p_{\mathcal D}(\mathbf{x})}\left[D_{\mathrm{KL}}\left(q_{\boldsymbol \phi}(\mathbf{z} | \mathbf{x}) \| p_{\boldsymbol \theta}(\mathbf{z} )\right) \right]$, the encoder  optimized in such a way will target to make the mean of all mappings close to zero and whiten the ellipsoids.   

Without a special guidance, these mappings will converge to the zero mean unit variance Gaussian marginal shape under asymptomatically many input mappings. Obviously, there is a little control on this process but the final goal of stochastic minimization of the upper bound on the mutual information $ I_{\boldsymbol \phi}({\bf X} ;{\bf  Z})$ considered as a "compression" is achieved according to the IB framework.

  {\bf $\beta$-VAE   \cite{higgins2017beta}}  is linked to (\ref{BIBN_unsupervised}) in the same way as the VAE but with a varying  relaxation parameter $\beta$:
  \begin{equation}
\label{b-VAE}
\mathcal{L}_{\beta-\mathrm{VAE}}(\boldsymbol \theta, \boldsymbol \phi) = \mathbb{E}_{p_{\mathcal D}({\bf x})}\left[  D_{\mathrm{KL}}(q_{\boldsymbol \phi}(\mathbf{z}|{\bf X=x} ) \| p_{\boldsymbol \theta}(\mathbf{z}))\right] -  \beta\mathbb{E}_{p_{\mathcal D}({\bf x})}\left[\mathbb{E}_{q_{\boldsymbol \phi}({\bf z} | {\bf x})}\left[\log p_{\boldsymbol \theta}({\bf x} | {\bf z})\right]\right].
 \end{equation}

The main advantage of  $\beta$-VAE  over VAE is a possibility to relax the described above stochastic "compression" via mapping everything to a big Gaussian "heap"  by applying the relaxation parameter $\beta$ that might give more preference to the reconstruction cost. By increasing $\beta$, one might achieve a sort of "disentangliation", yet weakly controllable by one global parameter, by allowing Gaussian clouds in the latent space to be far away from each other by less satisfying the KL-term constraint to fit the marginally Gaussian distribution. The semantically similar inputs might be mapped closer thus creating a sort of clusters that might be interpreted as a disentangled representation. Surely, it is only an interpretation of such a relaxed stochastic mapping and the process of "semantic clustering" highly depends on statistics of data. It seems to be quite difficult to achieve a semantically meaningful encoding and interepretability of the latent space without either at least some weak supervision or specially constructed latent space. 
%We will comment on this option below.

%  We need to find a bound on $I_{\boldsymbol \phi}({\bf X} ;{\bf  Z}) $ via $\mathbb{E}_{p_{\mathcal D}({\bf x})}\left[\mathbb{E}_{q_{\boldsymbol \phi}({\bf z} | {\bf x})}\left[\log p_{\boldsymbol \theta}({\bf x} | {\bf z})\right]\right] $. The problem is that we have a sum of two non-negative terms. (Similar to DP INQ: the mutual information on the left is larger that any of two additive components since they are non-negative). It means that $ I_{\boldsymbol \theta, \boldsymbol \phi}({\bf Z}; {\bf  X} )   \geq  \mathbb{E}_{p_{\mathcal D}({\bf x})}\left[\mathbb{E}_{q_{\boldsymbol \phi}({\bf z} | {\bf x})}\left[\log p_{\boldsymbol \theta}({\bf x} | {\bf z})\right]\right] $.
%  
 % Similarly, the second term is also upper bounded as $ I_{\boldsymbol \theta, \boldsymbol \phi}({\bf Z}; {\bf  X} )   \leq  \mathbb{E}_{p_{\mathcal D}({\bf x})}\left[\mathbb{E}_{q_{\boldsymbol \phi}({\bf z} | {\bf x})}\left[\log p_{\boldsymbol \theta}({\bf x} | {\bf z})\right]\right] $,  since $D_{\mathrm{KL}}\left(p_{\mathcal D}(\mathbf{x} ) \| p_{\boldsymbol \theta}(\mathbf{x})\right) \geq 0$.

%{\em Something important according to the Japanese paper that according to the used parametrization in the VAE}: the VAE tries to map ALL data samples points to the same $\bf z$ - a sort of clustering. This minimizes the mutual information  $ I_{\boldsymbol \phi}({\bf X} ;{\bf  Z})$ but it kills all structure in data. Instead, the Japanese paper states that they only map two points ${\bf x}_i$ and  ${\bf x}_j$ to a common $\bf z$ at once! We should investigate it more. 

{\bf AAE  \cite{makhzani2015adversarial}} can be defined according to the equivalent Lagrangian cost: 
\begin{equation}
\label{AAE}
\mathcal{L}_{\mathrm{AAE}}(\boldsymbol \theta, \boldsymbol \phi) =  D_{\mathrm{KL}}(q_{\boldsymbol \phi}(\mathbf{z} ) \| p_{\boldsymbol \theta}(\mathbf{z})) - \beta \mathbb{E}_{p_{\mathcal D}({\bf x})}\left[\mathbb{E}_{q_{\boldsymbol \phi}({\bf z} | {\bf x})}\left[\log p_{\boldsymbol \theta}({\bf x} | {\bf z})\right]\right],
 \end{equation}
where we do not explicitly consider the technical details of KL-divergence approximation and computation whereas one can use adversarial discriminator for this purpose or the maximum mean discrepancy (MMD) \cite{gretton2012kernel} based discriminator.
 
It should be pointed out that (\ref{AAE}) contains the term (C) which origin can be explained in the same way as for the VAE. Despite of the fact that the term (B)  indeed appears in (\ref{AAE}) with the opposite sign, it cannot be interpreted either as an upper bound on $I_{\boldsymbol \phi}({\bf X} ;{\bf  Z})$ similarly to the VAE or as a lower bound. The goal of AAE is to minimize the reconstruction loss or to maximize the log-likelihood by ensuring that the latent space marginal distribution $q_{\boldsymbol \phi} ({\bf z}) $ matches the prior $p_{\boldsymbol \theta}({\z})$. The latter  corresponds to the minimization of $D_{\mathrm{KL}}\left(q_{\boldsymbol \phi} ({\bf z}) \| p_{\boldsymbol \theta}(\mathbf{z})\right)$.

It is interesting to point out that the original AAE paper considers as a potential encoding all options that include: a {\em deterministic encoding}, i.e., ${\bf z} = f_{\boldsymbol \phi}({\x})$, as well as the considered in section 3 {\em stochastic encodings}. A nice flexibility of AAE comes from a possibility to match the observed marginal distribution $\qz$ to a desired targeted distribution $
p_{\boldsymbol \theta}({\z})$ without the need to have  explicitly defined distributions in contrast to the VAE. 

%\todo{To develop the first MI for the deterministic encoding.}

An actual implementation of AAE is based on the deterministic encoding. We can imagine this sort of mapping  by considering Figure \ref{Fig:4}. A point of the observation space is mapped just to one point in the latent space. Under the deterministic encoder the mutual information  $I_{\boldsymbol \phi}({\bf X} ;{\bf  Z}) =  H_{\boldsymbol \phi}({\bf Z})$ since $ H_{\boldsymbol \phi}({\bf Z}|{\bf X}) = 0$\footnote{One can use a variational decomposition $H_{\boldsymbol \phi}({\bf Z})  = - \mathbb{E}_{ q_{\boldsymbol \phi}(\mathbf{z}) } \left[\log  q_{\boldsymbol \phi}(\mathbf{z})\frac{p_{\boldsymbol \theta}({\z})}{p_{\boldsymbol \theta}({\z})} \right] = H(q_{\boldsymbol \phi}(\mathbf{z});p_{\boldsymbol \theta}(\mathbf{z})) - D_{\mathrm{KL}}(q_{\boldsymbol \phi}(\mathbf{z} ) \| p_{\boldsymbol \theta}(\mathbf{z}))$. Thus, if one wants to reduce the entropy of latent space to the entropy $H_{\boldsymbol \theta}({\bf Z})$ of targeted distribution $ p_{\boldsymbol \theta}(\mathbf{z})$, one should ensure that the encoder targets $q_{\boldsymbol \phi} ({\bf z}) \rightarrow p_{\boldsymbol {\theta}}({\bf z})$  leading to   $H(q_{\boldsymbol \phi} ({\bf z}); p_{\boldsymbol {\theta}}({\bf z})) \rightarrow H_{\boldsymbol \theta}({\bf Z})$. Therefore, the term (B) in the AAE follows from the minimization of  $D_{\mathrm{KL}}\left(q_{\boldsymbol \phi} ({\bf z}) \| p_{\boldsymbol \theta}(\mathbf{z})\right)$. }. 

%\todo{To finalize above AAE linl to KL.}

That is why the ability to compress the observation space to the latent space or to generate from the latent space comes from the relationship between the entropy of observation space distribution $\pdx$ and targeted latent space distribution  $p_{\boldsymbol \theta}({\z})$. If the entropy of the observation space is large, i.e., the data are on a complex distributed manifold with a large variance, and the latent space is characterized by a small variance, many samples from the observation space will be mapped very closely to meet the KL-term constraint on the marginal latent space distribution. Naturally, it is a form of "deterministic" compression leading to the reduction of entropy by a "collusion" of many samples from the observation space in the latent space. It should be noticed that in this case, the centroids typically used in quantization based compression are not even used. At the same time, the "continuity" of latent space filling is determined by the randomness of $\pdx$ with respect to $p_{\boldsymbol \theta}({\z})$. If for some reason $p_{\boldsymbol \theta}({\z})$ is chosen to be relatively "broad",  it is not excluded that one might observe some "holes" in the latent space as a result of such a mapping.

 Nevertheless, as shown in Figure \ref{Fig:3}, one can impose any constraint on  $p_{\boldsymbol \theta}({\z})$ like Gaussian, Laplacian or even sparsifying prior. Moreover, one can predefine some centroids or clusters and target that the closest samples in the observation space to be mapped into the same centroids. In this sense, the AAE can also implement a form of deterministic compression by clustering. 
 
 At the same time, one can relax the quantization requirement to map an input to exactly one closest centroid and instead to envision some relaxation within the allowed KL-term. These options are not directly implemented in the AAE but can be envisioned. We mention and consider them in view of a link to InfoVAE and generative compression that will be addressed in the next section.

{\bf InfoVAE \cite{zhao2017infovae}} consists of 3 terms obtained by adding the regularisation term $ I_{\boldsymbol \phi}({\bf X} ;{\bf  Z}) $ to an alternative form of the VAE. Since this original way of deriving InfoVAE is not straightforward and does not naturally comes from the IB framework, we will show that the InfoVAE has its BIB-AE counterpart with the terms (A), (B) and (C) and can be defined according to the Lagrangian\footnote{The original InvoVAE contains different multipliers in front of KL-terms.}:
\begin{equation}
 \begin{aligned} 
\label{InfoVAE}
\mathcal{L}_{\mathrm{InfoVAE}}(\boldsymbol \theta, \boldsymbol \phi) & =\mathbb{E}_{p_{\mathcal D}(\mathbf{x})}\left[D_{\mathrm{KL}}\left(q_{\boldsymbol \phi}(\mathbf{z} | \mathbf{X=x}) \| p_{\boldsymbol \theta}(\mathbf{z} )\right) \right] -  D_{\mathrm{KL}}(q_{\boldsymbol \phi}(\mathbf{z} ) \| p_{\boldsymbol \theta}(\mathbf{z})) 
 \\ & - \beta \mathbb{E}_{p_{\mathcal D}({\bf x})}\left[\mathbb{E}_{q_{\boldsymbol \phi}({\bf z} | {\bf x})}\left[\log p_{\boldsymbol \theta}({\bf x} | {\bf z})\right]\right].
 \end{aligned} 
 \end{equation}

In fact,  the terms (A) and (B) correspond to $I_{\boldsymbol \phi}({\bf X} ;{\bf  Z}) $ while the term (C) corresponds to $ I^{\text{VAE}} _{\boldsymbol \theta, \boldsymbol \phi}({\bf Z}; {\bf  X} )$. Besides, it should also be pointed out that in the original paper \cite{zhao2017infovae} the above three terms have not been used jointly in the reported simulations.  Instead, the original InfoVAE uses 2 terms depending on the VAE form, i.e., the terms (A) and (C), or the terms (B) and (C), i.e., the AAE form.

The InfoVAE can also be considered as yet another form of compression by the minimization of $I_{\boldsymbol \phi}({\bf X} ;{\bf  Z})$. Since it contains both KL-terms  (A) and (B) in  $I_{\boldsymbol \phi}({\bf X} ;{\bf  Z})$, the encoder can minimize  $I_{\boldsymbol \phi}({\bf X} ;{\bf  Z})$ by seeking an equality between the terms (A) and (B) since both terms are non-negative. One can consider the presence of term (B) with the regularization parameter $\beta$ as a regularization of VAE term (A). As a result, it will relax the condition to map all conditional distributions to one Gaussian heap how it is done in the VAE case.

Having considered all these connections, it should be pointed out that the interpretability of the latent space in all considered methods is a quite complex task unless special supervised constraints are imposed how it was finally suggested in a semi-supervised AAE framework. For this reason, we will also consider other possibilities of controllable latent space encoding and generation using generative compression. However, it should be noted that the initial goal of this type of encoding has different roots and requires the selection of optimal distribution to meet a rate-distortion trade-off.

{\bf GANs \cite{goodfellow2014generative}}: not pretending to consider the whole GAN family, we can mention that the IB considered for  the generative adversarial models in section \ref{IB_generative} makes it possible to link GAN with BIB-AE. Considering the generation from the targeted latent space distribution $p_{\boldsymbol  \theta}({\z})$ via the generator $\pthx_z$ one uses  (\ref{IB_G:lower_bound}) that corresponds to the terms (D) and (C) in (\ref{MI_decoder_sum}), respectively. Therefore, the BIB-AE is linked to GANs via the IB framework. 

It should be remarked that the original GAN does not include the likelihood term (C). However, according to the BIB-AE analysis, this regularizer naturally follows from the IB framework. It is interesting to mention that Rosca {\em et. al.} \cite{rosca2017variational} have considered this option as a potential solution to the GAN mode collapse problem.

{\bf VAE/GAN \cite{larsen2015autoencoding}}: an option to use jointly the VAE represented by term (A) and (C) and the GAN represented by term (D) was envisioned in VAE/GAN model. An equivalent VAE/GAN Lagrangian is formulated as: 
\begin{equation}
 \begin{aligned} 
\label{VAE_GAN}
 \mathcal{L}_{\mathrm{VAE/GAN}}(\boldsymbol \theta, \boldsymbol \phi) & = \mathbb{E}_{p_{\mathcal D}({\bf x})}\left[  D_{\mathrm{KL}}(q_{\boldsymbol \phi}(\mathbf{z}|{\bf X=x} ) \| p_{\boldsymbol \theta}(\mathbf{z}))\right] - \beta \mathbb{E}_{p_{\mathcal D}({\bf x})}\left[\mathbb{E}_{q_{\boldsymbol \phi}({\bf z} | {\bf x})}\left[\log p_{\boldsymbol \theta}({\bf x} | {\bf z})\right]\right]  \\ & + \beta D_{\mathrm{KL}}\left(p_{\mathcal D}(\mathbf{x} ) \| p_{\boldsymbol \theta}(\mathbf{x})\right).
 \end{aligned} 
 \end{equation} 
In the original paper, the log-likelihood term was replaced by a special metric in the latent space\footnote{One can use both encoded-reconstructed samples and samples generated from $p_{\boldsymbol \theta}({\z})$ in the third term for the adversarial discrimination.}.

In conclusion, many existing variations of VAE and GAN families can be considered directly from the BIB-AE framework perspectives. The main difference between these approaches, where either the VAE based on ELBO or GAN are taken as a basis and then some regularization terms are added, and the proposed one is in a fact that we proceed directly with the IB formulation and impose the corresponding bounds on the mutual information components of the IB.

Extending the same methodology, next we consider a compression formulation of IB from the Shannon's rate-distortion perspectives and link it with generative models.

\vspace{-3mm}
\subsection{Shannon's rate-distortion and generative compression AEs}
\vspace{-3mm}
In the previous analysis, the targeted latent space distribution was assumed to be any manifold specified by $p_{\boldsymbol \theta}({\bf z})$.  However, if one wants additionally to have a latent space with a bounded rate below the entropy $H_{\boldsymbol \theta}({\Z}) = - \mathbb{E}_{p_{\boldsymbol \theta}({\bf z})}[\log p_{\boldsymbol \theta}({\bf z})]$, i.e., targeting some compression, yet providing the best reconstruction and possibly generation from the latent space samples, it is of interest to link the considered analysis to the Shannon's rate-distortion theory.

Since the latent space of compression AE should be limited to some rate $R_Q$, we will assume that the latent space consists of a codebook $\mathcal{C} = \{ {\bf c}_1, {\bf c}_2, \cdots, {\bf c}_L \}$, containing the codewords ${\bf c}_i \in {\mathbb R}^{n_z}$ of dimension $n_z$ with probabilities $\{ p_j\}_{j=1}^L$ such that $R_Q=- \sum_{j=1}^Lp_j \log p_j$. The codewords of $\mathcal{C} $ can be considered as realizations  or centroids generated from $p_{\boldsymbol \theta}(\bf z)$ that makes it conceptually similar to the AAE. This is conceptually shown in Figure \ref{Fig:3} as "compressed" latent space. 

At the same time, an essential simplification comes from the fact that the encoder is deterministic and  maps the input to one of the above centroids. This can be achieved by a vector quantizer ${ \bf \hat z} = Q(f_{\boldsymbol \phi} ({\bf x})) := \argmin_{1 \leq j \leq L} ||  f_{\boldsymbol \phi} ({\bf x)} - {\bf c}_j  ||_2$, where $f_{\boldsymbol \phi} (\bf x)$ denotes a deterministic encoder and $Q(.)$ a vector quantizer (VQ). Hence, the distribution of the quantized latent space is $p_{\boldsymbol \theta}({\bf \hat z})=  \sum_{j=1}^L p_j{\boldsymbol \delta}({ \bf \hat z} - {\bf c}_j)$ that defines the rate $R_Q$.

{\bf Shannon's rate-distortion \cite{cover2012elements}} can be expressed as a special case of (\ref{BIBN_unsupervised}) with  (\ref{MI_encoder_sum}) and (\ref{MI_decoder_sum}):
\begin{equation}
\label{Shannon}
\mathcal{L}_{\mathrm{Shannon-AE}}(\boldsymbol \theta, \boldsymbol \phi) =  I^Q_{\boldsymbol  \phi}({\bf  X} ; {\bf \hat Z}) -\beta   I^{\mathrm{U}_L}_{\boldsymbol \theta, \boldsymbol \phi}({\bf \hat Z}; {\bf  X} ).
 \end{equation}
It is easy to show that $I_{\boldsymbol  \phi}({\bf  X} ; {\bf \hat Z}) = I^Q_{\boldsymbol  \phi}({\bf  X} ; {\bf \hat Z}) = H_{\boldsymbol \phi}({\bf \hat Z})$ due to the deterministic encoding with quantization, while $ I^{\mathrm{U}_L}_{\boldsymbol \theta, \boldsymbol \phi}({\bf \hat Z}; {\bf  X} )$ is reduced to the term (C) that under the deterministic decoding further reduces to $ \mathbb{E}_{p_{\mathcal D}({\bf x})}\left[\log p_{\boldsymbol \theta}({\bf x} | {\bf \hat z})\right]$. This term corresponds to the reconstruction distortion that is often expressed as the $\ell_2$-norm that in turns corresponds to the Shannon's lower bound on rate-distortion function. Therefore, the classical compression schemes satisfy the trade-off between the rate $I^Q_{\boldsymbol  \phi}({\bf  X} ; {\bf \hat Z}) = R_Q$  and distortion $ \mathbb{E}_{p_{\mathcal D}({\bf x})}\left[\log (- p_{\boldsymbol \theta}({\bf x} | {\bf \hat z}))\right] = D$. Finally, the latent space distribution $p_{\boldsymbol \theta}({\bf \hat z})$ is optimized to ensure the achievability of rate-distortion limit. This is a fundamental difference with the AAE, where the latent space distribution is chosen in advance for the technical reasons.

It is important to note that the Shannon's rate distortion framework in the considered interpretation is closely linked with the AAE, when the targeted distribution latent space is represented by the compression codebook Figure \ref{Fig:4}. The difference is in the practical implementation. The VQ implementation assumes a hard assignment of the input to one of centroids\footnote{The multiple assignments are also possible that is know as {\em soft-encoding}.}, whereas the AAE proceeds with the optimization of the KL-term to fit the targeted latent space distribution. It means that some deviation from the centroids is still possible. However, in both cases to proceed with the generation from the latent compressed space, one needs to ensure a proper randomness. Otherwise, the space of reconstructed signals will correspond to the number of centroids in the latent space. For this reason, we will consider a generative compression and link it with the IB framework.

%\todo{To explain why we talk about it: since we envision deterministic encoding similar to AAE but to avoid holes and to have more powerful generative capacity we envision and add noise to the centroids before passing it to the decoder.}

{\bf Generative compression \cite{agustsson2018generative, santurkar2018generative, tschannen2018deep, blau2019rethinking}} can be considered as an "extension" of Shannon's rate distortion with the Lagrangian:
\begin{equation}
\label{CG_AE}
\mathcal{L}_{\mathrm{GC-AE}}(\boldsymbol \theta, \boldsymbol \phi) =  I^Q_{\boldsymbol  \phi}({\bf  X} ; {\bf \hat Z}) -\beta   I^{ \mathrm{U}_L}_{\boldsymbol \theta, \boldsymbol  \phi}({\bf \hat Z} + {\bf U}; {\bf  X}).
 \end{equation}
The first term is the same as in the classical compression setup, while the second one contains a stochastic component achieved by the addition of the permutation ${\bf U} \sim p_{\bf u}({\bf u})$ to the centroids\footnote{We use another variable $\bf u$ for the randomization of centroids to reflect a fact that it is assigned to the decoder part in contrast to the randomization based on the encoder randomization using $\boldsymbol \epsilon$.}. At the same time, it contains both equivalent terms (C) and (D) in (\ref{MI_decoder_sum}). In practice, the KL-divergence is lower bounded by $f$-divergence that is implemented in a form of adversarial loss based on a density ratio estimation \cite{nowozin2016, mohamed2016learning} or its Wasserstein's counterpart \cite{arjovsky2017wasserstein}. In the original generative compression papers, the origin of the $f$-divergence term interpreted as a perceptual loss was only explained from the heuristic point of view to make highly compressed fragments of images under a low compression rate to look more naturally but not necessarily to be close to the original fragments. However, we can trace the origin of this term as an outcome of the IB factorization. %Moreover, if there is no permutation term $ \bf u$ and the perceptual loss (D), the generative compression naturally reduces to the Shannon's rate distortion setup. 

\vspace{-3mm}
\subsection{Novelty detection AEs}
\vspace{-2.5mm}
The novelty detection problem aims at detecting outliers with respect to some manifold represented by the training data set. It assumed that similarly to the unsupervised setup, the training set consisting of $N$ samples is given. One can use different techniques to measure the relevance of a test sample to the training set or even to train a one class classifier for this purpose.

Alternatively, one can consider a novelty detection problem from the position of unsupervised IB framework in the BIB-AE formulation. It is interesting to note that \cite{pidhorskyi2018generative} proposed the architecture similar to the BIB-AE presented in Figure \ref{Fig:set_up} and trained with the terms (B), (C) and (D) for the novelty detection. The AE trained in this way might use several metrics such as output of term (D) to detect outliers. This also corresponds to a one-class classification problem. Therefore, the mechanism of novelty detection can be seen from the perspective of using the BIB-AE architecture.

\vspace{-2mm}
\section{Conclusions}
\vspace{-2mm}
In this paper, we considered the IB for several practical tasks covering supervised, unsupervised, generative adversarial, generative compressive and novelty detection models. We show that the IB for all these models reduces to four terms in the Lagrangian cost. We call this formulation as BIB-AE. This formulation is closely linked with many  models ranging from the VAE to VAE/GAN.

Besides this remarkable similarity, we note that this connection is seen via the IB framework with application of variational approach to the decomposition of mutual information terms in contrast to the VAE family that is based on various attempts to regularize the ELBO. As a result, the interpretability of obtained results and connection between methods leads to different conclusions. 

Along the same line, we consider the new framework of generative compression in a close link to the IB  framework whereas the original works on the generative compression considered it from the "perceptual" perspectives by adding the regularizer similar to the ELBO.

Finally, we also show that the novelty detection problem in the recent interpretability of AE encoding with the adversarial loss can be linked to the BIB-AE interpretation. Altogether the performed analysis gives new insights on the connections between different problems and methods and creates an interesting basis for the interpretability of the latent space.

%The introduced BIB-AE might be of interest in several applications addressing generative models, compression and novelty detection.

{\bf Acknowledgement}

The research was supported by the SNF project No. 200021-182063. The authors are thankful to Behrooz Razeghi for his feedback and fruitful discussions.

\newpage

%%%% Appendix A %%%%%%
% To copy
{\bf Appendix A}

In this part, we derive a lower bound on $I({\bf Z};{\bf C})$. According to the definition (\ref{MI_2nd_unsupervised_def1}), this mutual information can be further decomposed as:
\begin{equation}
 \begin{aligned} 
\label{MI_2nd_unsupervised_def_copy}
 I({\bf Z} ;{\bf  C})  & =   \mathbb{E}_{\pcz}  \left[  \log \frac{\pc_z}{\pdc}  \right]
  \\ &  = -  \mathbb{E}_{p({\bf c})}  \left[  \log p({\bf c})\right] + \mathbb{E}_{\pcz}  \left[  \log p(\mathbf{c}| \mathbf{z})\right].
 \end{aligned} 
\end{equation}
The first term of this decomposition corresponds to the entropy of classes $H({\bf C})  = - \mathbb{E}_{p({\bf c})}  \left[ \log p({\bf c}) \right]$.

We consider the second term since the transition probability $p(\mathbf{c}| \mathbf{z})$ is unknown. At the same time, it can be written as:
\begin{equation}
 \begin{aligned} 
\label{MI_2nd_unsupervised_cond_entropy1}
\mathbb{E}_{\pcz}  \left[  \log p(\mathbf{c}| \mathbf{z})\right] & =   \int_{\bf c}\int_{\bf z}  p({\bf c,z})   \log p(\mathbf{c}| \mathbf{z})   \ \mathrm{d}{\bf c} \ \mathrm{d}{\bf z}.
 \end{aligned} 
\end{equation}
The expectation is with respect to the joint distribution $\pcz$ that can also be defined via the marginalization $\pcz = \int_{\bf x} p({\bf c,z,x})d{\bf x} = \int_{\bf x} p({\bf c,x})\qzx \mathrm{d}{\bf x}$. Therefore, combing these results, one can obtain:    
\begin{equation}
 \begin{aligned} 
\label{MI_2nd_unsupervised_cond_entropy2}
\mathbb{E}_{\pcz}  \left[  \log p(\mathbf{c}| \mathbf{z})\right] & =   \int_{\bf c}\int_{\bf z}\int_{\bf x}  p({\bf c,x})\qzx  \log p(\mathbf{c}| \mathbf{z})    \ \mathrm{d}{\bf c}  \ \mathrm{d}{\bf z} \ \mathrm{d}{\bf x}
 \\ &  = \mathbb{E}_{p({\bf c},{\bf x} )}  \left[   \mathbb{E}_{q_{\boldsymbol \phi}({\bf z}|{\bf x})} \left[  \log p(\mathbf{c}| \mathbf{z}) \right]\right].
 \end{aligned}
\end{equation}
To overcome the problem of unknown $p(\mathbf{c}| \mathbf{z})$, we will apply a variational distribution $p_{\boldsymbol \theta}(\mathbf{c}| \mathbf{z})$ parametrized via a set of parameters $\boldsymbol \theta$ to approximate $p(\mathbf{c}| \mathbf{z})$. This can be considered as a bypass network and formulated as:
\begin{equation}
 \begin{aligned} 
\label{MI_2nd_unsupervised_cond_entropy3}
 \mathbb{E}_{\pcz}  \left[  \log p(\mathbf{c}| \mathbf{z})\right]  & =   \mathbb{E}_{p({\bf c},{\bf x} )}  \left[   \mathbb{E}_{q_{\boldsymbol \phi}({\bf z}|{\bf x})} \left[  \log p(\mathbf{c}| \mathbf{z}) \frac{ p_{\boldsymbol \theta}(\mathbf{c}| \mathbf{z})}{p_{\boldsymbol \theta}(\mathbf{c}| \mathbf{z})} \right]\right]
  \\ &  = \mathbb{E}_{p({\bf c},{\bf x} )}  \left[   \mathbb{E}_{q_{\boldsymbol \phi}({\bf z}|{\bf x})} \left[  \log p_{\boldsymbol \theta}(\mathbf{c}| \mathbf{z}) \right]\right]
 +   \mathbb{E}_{p({\bf c},{\bf z} )}  \left[ \log  \frac{p(\mathbf{c}| \mathbf{z})}{p_{\boldsymbol \theta}(\mathbf{c}| \mathbf{z})} \right],
 \end{aligned}
\end{equation}
where in the second term we used the expectation defined in (\ref{MI_2nd_unsupervised_cond_entropy2}).

At the same time, we can re-write $ p({\bf c},{\bf z} ) = p({\bf z})p( {\bf c}|{\bf z})$ that leads to\footnote{One can note that $p({\bf z}) = \qz$.}:
 \begin{equation}
  \begin{aligned} 
\label{MI_2nd_unsupervised_cond_entropy4}
 \mathbb{E}_{p({\bf c},{\bf z} )}  \left[ \log  \frac{p(\mathbf{c}| \mathbf{z})}{p_{\boldsymbol \theta}(\mathbf{c}| \mathbf{z})} \right] & = \mathbb{E}_{p({\bf z} )} \left[  \mathbb{E}_{p({\bf c}|{\bf z} )}   \left[ \log  \frac{p(\mathbf{c}| \mathbf{z})}{p_{\boldsymbol \theta}(\mathbf{c}| \mathbf{z})} \right]\right] \\ & =  \mathbb{E}_{p({\bf z} )}  \left[   D_{\mathrm{KL}}\left(p(\mathbf{c}|{\bf Z=z} ) \| p_{\boldsymbol \theta}(\mathbf{c}|{\bf Z=z})\right)  \right] = D_{\mathrm{KL}}\left(p(\mathbf{c}|{\bf z} ) \| p_{\boldsymbol \theta}(\mathbf{c}|{\bf z})\right).
  \end{aligned} 
\end{equation}

Since the KL-divergence $D_{\mathrm{KL}}\left(p(\mathbf{c}|{\bf z} ) \| p_{\boldsymbol \theta}(\mathbf{c}|{\bf z})\right) \geq 0$, we can lower bound (\ref{MI_2nd_unsupervised_cond_entropy3}) as:
 \begin{equation}
  \begin{aligned}
\label{MI_2nd_unsupervised_cond_entropy5}
 \mathbb{E}_{\pcz}  \left[  \log p(\mathbf{c}| \mathbf{z})\right]  & \geq \mathbb{E}_{p({\bf c},{\bf x} )}  \left[   \mathbb{E}_{q_{\boldsymbol \phi}({\bf z}|{\bf x})} \left[  \log p_{\boldsymbol \theta}(\mathbf{c}| \mathbf{z}) \right]\right].
 \end{aligned}
\end{equation}

Therefore, the mutual information (\ref{MI_2nd_unsupervised_def_copy}) can be lower bounded as  $I({\bf Z};{\bf C}) \geq I^{\mathrm{S}}_{\boldsymbol \theta, \phi}({\bf Z};{\bf C})$, where we define a lower bound as:
\begin{equation}
 \begin{aligned} 
\label{IB_S_lower_bound1}
I^{\mathrm{S}}_{\boldsymbol \theta, \boldsymbol  \phi}({\bf Z};{\bf C}) & \triangleq H({\bf C}) + \mathbb{E}_{p({\bf c},{\bf x} )}  \left[   \mathbb{E}_{q_{\boldsymbol \phi}({\bf z}|{\bf x})} \left[  \log p_{\boldsymbol \theta}(\mathbf{c}| \mathbf{z}) \right]\right]
 \\ & = H({\bf C}) - H_{\boldsymbol \theta, \boldsymbol \phi} ({\bf C}|{\bf Z}),
 \end{aligned} 
\end{equation}
where $H_{\boldsymbol \theta, \boldsymbol \phi} ({\bf C}|{\bf Z}) = - \mathbb{E}_{p({\bf c},{\bf x} )}  \left[   \mathbb{E}_{q_{\boldsymbol \phi}({\bf z}|{\bf x})} \left[  \log p_{\boldsymbol \theta}(\mathbf{c}| \mathbf{z}) \right]\right]$.

{\small
\bibliographystyle{unsrt}
\bibliography{egbib}
}

\end{document}